\documentclass[journal]{IEEEtran}
\usepackage{ulem}
\usepackage{graphicx}
\usepackage{bm}
\usepackage{varwidth}
\usepackage{algorithm}
\usepackage{algorithmic}
\usepackage{xspace}
\usepackage{fixfoot}
\usepackage[utf8]{inputenc} 
\usepackage[T1]{fontenc}    
\usepackage{hyperref}       
\usepackage{url}            
\usepackage{booktabs}       
\usepackage{amsfonts}       
\usepackage{nicefrac}       
\usepackage{microtype}      
\usepackage{xcolor}         
\usepackage{tikz}
\usepackage{calc}
\usepackage{cancel} 
\usepackage{verbatim}
\usepackage{graphicx}
\usepackage{rotating}
\usepackage{bbm}
\usepackage{latexsym}
\usepackage{times}
\usepackage{amssymb,amsmath,amsthm}
\usepackage{mathtools}
\usepackage{color}
\usepackage{makecell}
\usepackage{multirow}
\usepackage{epstopdf}
\usepackage{colortbl}
\usepackage{wrapfig}
\usepackage{pdflscape}
\usepackage{algorithm}
\usepackage[algo2e]{algorithm2e}
\usepackage{algorithmic}
\usepackage{enumitem}
\usepackage{tikz}
\usepackage{etoolbox}
\usepackage{booktabs}
\usepackage{xcolor}
\usepackage{subfigure}
\usepackage{diagbox}

\newtheorem{remark}{Remark}
\newtheorem{proposition}{Proposition}
\newtheorem{Proof}{Proof}

\usepackage{tablefootnote}
\usepackage{bm}
\usepackage{varwidth}
\newcommand*\circled[1]{\tikz[baseline=(char.base)]{
    \node[shape=circle,draw,inner sep=0.9pt] (char) {#1};}}
\definecolor{gray}{gray}{0.65} 
\usepackage{extarrows}

\ifCLASSINFOpdf
\else
\fi
\begin{document}
%
\title{Coarse-to-Fine Contrastive Learning on Graphs}
%
%
%

\author{Peiyao Zhao,
        Yuangang Pan,
        Xin Li*,~\IEEEmembership{Member,~IEEE,} 
        Xu Chen,
        Ivor W. Tsang,~\IEEEmembership{Fellow,~IEEE}
        and Lejian Liao
\thanks{The work of Peiyao Zhao and Xin Li was supported in part by NSFC under Grant 92270125 and 62276024. The work of Yuangang Pan was supported in part by A*STAR Career Development Fund (CDF) 2022 and in part by the A*STAR Centre for Frontier AI Research. The work of Ivor W. Tsang was supported in part by the A*STAR Centre for Frontier AI Research (CFAR) and in part by the Australian Research Council under Grant DP200101328. \textit{(Corresponding author: Xin Li.}) 

Peiyao Zhao, Xin Li, Lejian Liao are with the School of Computer Science \& Technology, Beijing Institute of Technology, Beijing, 100081 China (e-mail: peiyaozhao@bit.edu.cn; xinli@bit.edu.cn; liaolj@bit.edu.cn ).

Yuangang Pan is with the A*STAR Centre for Frontier AI Research, Singpore 138632 (e-mail: yuangang.pan@gmail.com).

Ivor W. Tsang is with the A*STAR Centre for Frontier AI Research,
Singapore 138632, and also with the Australian Artificial Intelligence Institute,
University of Technology Sydney, Ultimo, NSW 2007, Australia (e-mail: ivor\_tsang@ihpc.a-star.edu.sg).

Xu Chen is with Alibaba group (e-mail: xuchen2016@sjtu.edu.cn).
}


}


%
%

\markboth{Journal of \LaTeX\ Class Files,~Vol.~14, No.~8, August~2015}%
{Shell \MakeLowercase{\textit{et al.}}: Bare Demo of IEEEtran.cls for IEEE Journals}
%



\maketitle

\begin{abstract}
Inspired by the impressive success of contrastive learning (CL), a variety of graph augmentation strategies have been employed to learn node representations in a self-supervised manner. Existing methods construct the contrastive samples by adding perturbations to the graph structure or node attributes. Although impressive results are achieved, it is rather blind to the wealth of prior information assumed: with the increase of the perturbation degree applied on the original graph, 1) the similarity between the original graph and the generated augmented graph gradually decreases; 2) the discrimination between all nodes within each augmented view gradually increases. In this paper, we argue that both such prior information can be incorporated (differently) into the contrastive learning paradigm following our general ranking framework. In particular, we first interpret CL as a special case of learning to rank (L2R), which inspires us to leverage the ranking order among positive augmented views. Meanwhile, we introduce a self-ranking paradigm to ensure that the discriminative information among different nodes can be maintained and also be less altered to the perturbations of different degrees. Experiment results on various benchmark datasets verify the effectiveness of our algorithm compared with the supervised and unsupervised models.

\end{abstract}
\begin{IEEEkeywords}
Graph Representation Learning, Learning to Rank, Node Representation, Self-supervised Learning, Contrastive Learning.
\end{IEEEkeywords}
%
\IEEEpeerreviewmaketitle

\section{Introduction}
%
%
%
%
\IEEEPARstart{R}{ecent} years have seen a growing amount of interest in self-supervised learning (SSL), especially in efforts to develop generalized Graph Neural Networks (GNNs) via employing various mutual information estimators~\cite{DBLP:conf/icml/HassaniA20} for node classification~\cite{DBLP:conf/iclr/VelickovicFHLBH19, DBLP:conf/aaai/WanPY021, randomweights, haar} and graph classification~\cite{DBLP:conf/nips/YouCSCWS20, DBLP:journals/tnn/WuPCLZY21}. Contrastive learning (CL) is a popular technique of self-supervised learning. Existing methods~\cite{DBLP:conf/icml/HassaniA20,DBLP:conf/nips/YouCSCWS20,DBLP:conf/ijcai/JinZL00P21} based on CL have greatly benefited from multi-type augmented views generated by adding perturbations to graph structure and/or node attribute. Graph Contrastive Learning (GCL) aims to pre-train a generalized graph neural
network which can be effectively fine-tuned in downstream tasks. The GCL loss enforces the representation of an instance close to that in an augmented view (referred to as positive pairs) and far from that of other instances (referred to as negative pairs) during the pre-training phase. 

\begin{figure}[!t]\label{motivation}
    \centering
    \vskip-0.1in
    \includegraphics[width=0.42\textwidth]{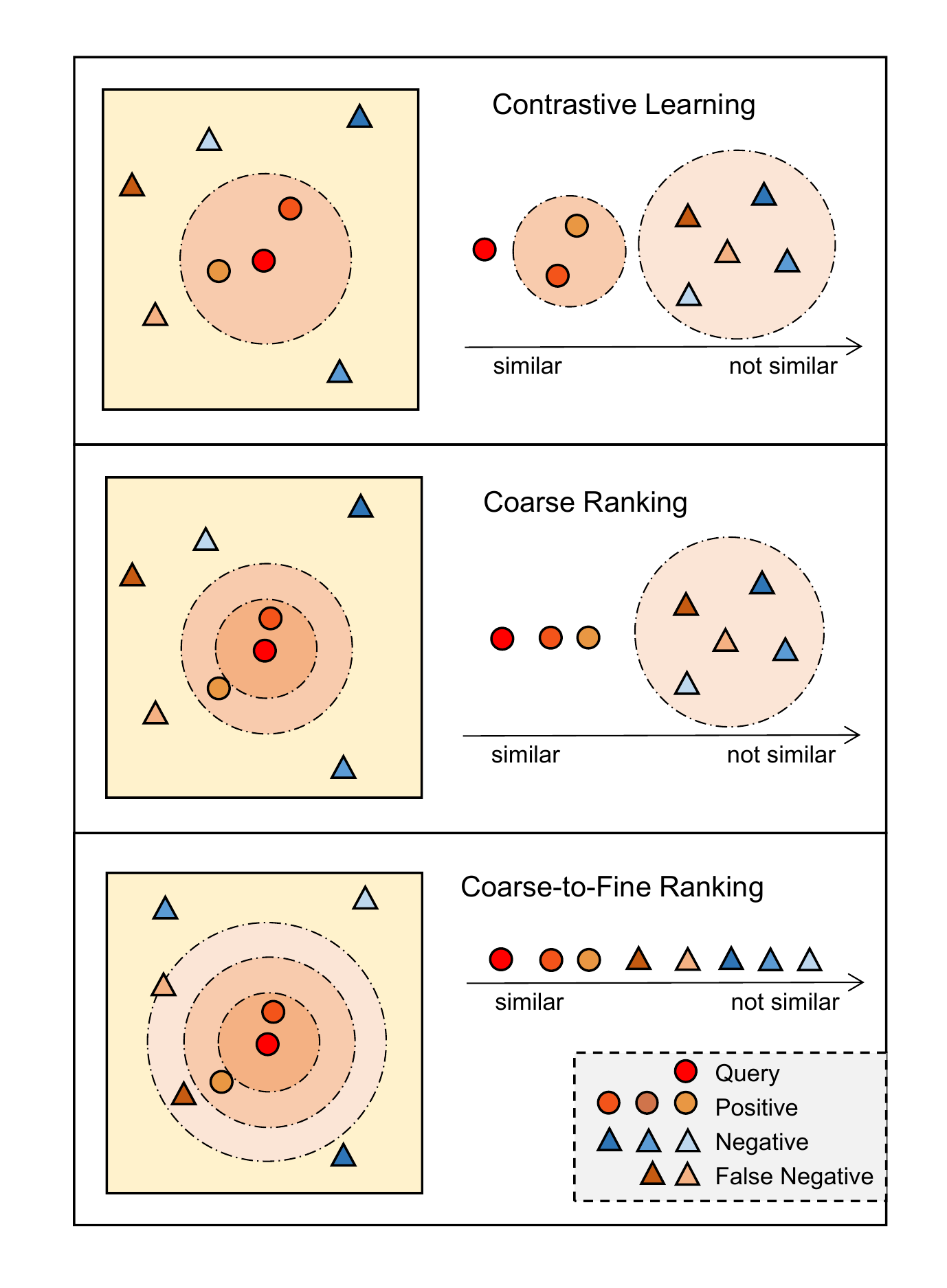}\vskip-0.2in
    \caption{Comparisons of different contrastive learning variants from the perspective of learning to rank. \textbf{Top:} Contrastive learning only discriminates between positive samples and negative samples.
    \textbf{Middle:} Our coarse ranking for CL can model the order relation among positive samples in two views of different perturbation degrees.
    \textbf{Bottom:} Our Coarse-to-Fine ranking for CL can simultaneously capture the discriminative information within positive samples and negative samples by ranking all samples within one learning to rank model.}\vskip-0.2in
\end{figure}

However, the traditional contrastive learning algorithms only model pair-wise views comparison indiscriminately and thus fail to distinguish instances across multiple augmented views
~\cite{DBLP:conf/kdd/QiuCDZYDWT20,DBLP:journals/corr/abs-2008-11416}. For example, a group of augmented views produced by adding different degrees of perturbation to the original graph can be ordered according to the perturbation degree (intuitively, the larger degree of the perturbation, the less similar the perturbed version is to the original graph). Most of self-supervised GNN models are unable to gain information conveyed in the ordered list. They can only train the encoder to discriminate between
positive samples from the joint distribution and negative samples from the marginal distribution, which fails to capture discrimination within all samples for enhancing representation learning (shown in Fig.\ref{motivation}).

To the best of our knowledge, multiple augmented views have never been explored to help enhance representation learning from a listwise ranking perspective. In this paper, we first theoretically prove that contrastive learning is a special case of learning to rank (L2R)~\cite{cao2007learning}, which naturally motivates us to leverage ranking to model the order relation implied by the different magnitudes of augmented views. We then propose a list-wise ranking loss for contrastive learning named coarse ranking, which incorporates the order relation among a variety of relevant representations in all augmented views, leading to more discriminative node representations. 
On this basis, we further design a fine-grained ranking loss to assign various pseudo-judgments for negative samples via a self-ranking mechanism, which allows negative samples to participate in sorting for further enhancing node representations. Eventually, we propose a general ranking loss for contrastive learning, Coarse-to-Fine contrastive learning (C2F), to unify the above proposed two ranking methods.

To be more specific, we first focus on incorporating the prior knowledge implied by the degrees of perturbations into contrastive learning. We adopt \textit{dropping edge} with varied rates as the perturbation strategy to generate correlated augmented views. We can observe from our experiments that the drop-edge ratio is a useful indicator of the similarity between graphs (depicted in APP.~\ref{insights}). Specifically, first, the average node similarity between one view and the original graph decreases with the increase of the drop-edge ratio of the view, which inspires us to sort positive samples based on the similarity between the query and positive samples in our coarse ranking module. 
Moreover, the average node similarity within one view decreases with the increase of the drop-edge ratio of the view. This phenomenon indicates that deleting a certain amount of connections in the graph may cause insufficient smoothness of node features, resulting in greater distances between nodes within the same class.
Thus we propose to incorporate negative samples into the ranking mechanism to supervise the fitting of augmented views. The supervision of negative samples can be achieved by assigning self-generated ground truths to them 
in our proposed fine-grained ranking model, which can distinguish among negative samples while preventing losing structure information.


To incorporate the above prior knowledge, we develop a general ranking loss to unify the above proposed two ranking methods, which assigns the self-generated ground truths to the prediction scores of all samples via a self-ranking method. Both the normalization of prediction scores and ground truth scores are element-wise calculated based on their score matrix, respectively. Fig.~\ref{motivation} shows the differences between CL, coarse ranking, and C2F ranking model. C2F can encourage the GNNs encoder to capture discriminative (order relation) information within various augmented views and negative samples, meanwhile inhibiting the loss of structure information caused by perturbation.
In short, we can summarize our main contributions as follows:
\begin{itemize}
 \item We analyze that contrastive learning is a special case of the learning-to-rank model and propose coarse ranking loss for CL to capture the order relation among positive views. The coarse contrastive learning naturally leverages the prior information implied by the degrees of perturbations to enhance node identity.
 \item We design a fine-grained ranking loss to assign ground truths for negative samples via a self-supervised ranking method. The fine-grained ranking module can prevent losing the graph structure information and enhance node features' smoothness.
\item We further design a general ranking framework for node representation learning via unifying the coarse and fine-grained ranking losses. Contrastive learning guided by ranking can capture discriminative information among positive and negative samples. 
 \item We evaluate the quality of the pre-trained encoder on several downstream classification benchmarks and ablate our framework. Empirical results demonstrate the superiority of the proposed method and the effectiveness of each component of our C2F.
 \end{itemize}
 
The layout of the paper is as follows. Section~\ref{related work} outlines the recent advancement of contrastive learning on graphs and clarifies the superiority of our proposed methods. Section~\ref{Background} describes the notations and the contrastive learning, and our reinterpretation of CL from the perspective of learning to rank. Section~\ref{Coarse-to-Fine Contrastive learning} introduces the proposed coarse-to-fine contrastive learning for a list of augmented graphs. Section~\ref{discuss} theoretically analyzes the self-generated ground truth matrix of our self-supervised ranking scheme in detail. In Section~\ref{Experiment}, the dataset and the experimental settings are detailed for the node classification task. The key experimental results are also summarized and the hyperparameters sensitivity regarding C2F is discussed. Section~\ref{Conclusion} draws conclusions and envisions the
 future work.

\section{Related work}~\label{related work}
Graph Neural Networks has gained increasing popularity in various domains due to its great expressive power and outstanding performance~\cite{DBLP:journals/tnn/WuPCLZY21,bacciu2022explaining,zou2022exploring}. Contrastive learning, learning to be invariant to different data
augmentations has become an important research subject in self-supervised representation learning recently~\cite{DBLP:journals/corr/abs-2006-08218}~\cite{ malkinski2022multi, liu2022contrastive,yang2022ncgnn}. With the advancement of CL, self-supervised graph representation learning has been improved significantly, and has been applied in various downstream tasks, such as node-level classification~\cite{DBLP:conf/iclr/VelickovicFHLBH19,DBLP:conf/icml/HassaniA20, DBLP:conf/kdd/QiuCDZYDWT20, DBLP:journals/corr/abs-2008-11416}, graph-level classification~\cite{DBLP:conf/nips/YouCSCWS20,DBLP:conf/iclr/SunHV020} and graph clustering~\cite{DBLP:conf/ijcai/ZhaoYWYD21}~\cite{li2022self}, neuronal morphological analysis~\cite{zhao2022graph}, anomaly detection~\cite{liu2021anomaly}. In this paper, we focus on the pretraining of graph neural networks to learn node-level representations favoring the node classification task.

Graph contrastive learning methods utilize the data augmentation strategy to generate augmented views and draw negative or positive samples from the views. Deep Graph Infomax (DGI)~\cite{DBLP:conf/iclr/VelickovicFHLBH19} constructed a negative view via row-wise shuffling attribute matrix and then maximized the mutual information between nodes representations (negative samples) randomly drawn from the negative view, and the global representations of the original graph (the positive sample). InfoGraph~\cite{DBLP:conf/iclr/SunHV020} followed the architectures of DGI to obtain graph representations for the downstream graph-level classification task. 

Graph Contrastive Coding (GCC)~\cite{DBLP:conf/kdd/QiuCDZYDWT20} extended MOCO~\cite{DBLP:conf/cvpr/He0WXG20} to graphs, which discriminated subgraphs sampled for a certain vertex and subgraphs sampled for other vertices for the pre-training GNN encoder. Graph Contrastive Learning (GraphCL)~\cite{DBLP:conf/nips/YouCSCWS20} compared four types of graph augmentations incorporating various priors and minimized the agreement between other graphs and the augmented graph to obtain graph representations applied to graph classification. Contrastive multi-view representation learning (MVGRL)~\cite{DBLP:conf/icml/HassaniA20} compared different combinations of augmentation strategies to discuss the effect of various perturbations. Graph Contrastive learning with Adaptive augmentation (GCA)~\cite{DBLP:conf/www/0001XYLWW21} designed an adaptive manner to corrupt the original graph according to graph structure and attributes. GMI~\cite{DBLP:conf/www/PengHLZRXH20} measured the correlation between input graphs and high-level hidden representations from two aspects of node features and topological structure. ~\cite{zhao2022graph} proposed a morphology-aware contrastive graph neural network for large-scale neuronal morphological representation learning.~\cite{tang2022contrastive} proposed an interpretable hierarchical signed graph representation learning model to extract graph representation from brain functional networks.~\cite{he2022analyzing} unified attribute completion and representation learning in the proposed unsupervised heterogeneous graph contrastive learning framework.~\cite{DBLP:journals/tnn/LiuLPGZK22} fully exploited the local information by sampling a type of contrastive instance pair and proposed a well-designed graph neural network-based contrastive learning model to learn informative embedding from high-dimensional attributes and local structure.

In computer vision, a supervised contrastive-based method~\cite{frosst2019analyzing} is proposed to utilize the supervised label information to select positive samples, namely, using other images in the same class as the positive samples for the given query, which incorporated multiple positive samples. Contrastive Multiview Coding (CMC)~\cite{tian2020contrastive} can scale to any number of views by maximizing mutual information between multiple views of the same scene, which can capture underlying scene semantics.~\cite{malkinski2022multi} treated Raven's progressive matrices into the multilabel classification, and then proposed a generalization of the noise contrastive estimation algorithm to the cases of multilabel samples and a new sparse rule encoding scheme. Liu et al.~\cite{liu2022contrastive} developed a clustering-based contrastive self-supervised learning model to capture the structure views and scenes which mapped SAR images from pixel space to high-level embedding space and facilitated the node representations and message passing.~\cite{wang2022graph} proposed a self-supervised graph-based contrastive learning framework, where the well-designed GNN is capable of mining the local transformational invariance and global textual knowledge.

Recent graph CL approaches designed specific network structures to dismiss negative samples and developed several techniques to prevent trivial solutions (e.g., gradient stopping, momentum encoder, predictor network). Bootstrapped Graphs Latents (BGRL)~\cite{thakoor2021bootstrapped} extended the outstanding CL method on images, BYOL~\cite{DBLP:conf/nips/GrillSATRBDPGAP20}, to graphs, which learned node representations by training an online encoder to predict the representation of a target one, without using negative pairs. G-BT~\cite{DBLP:journals/corr/abs-2106-02466} extended Barlow Twins~\cite{DBLP:conf/icml/ZbontarJMLD21} on images to graphs for utilizing a cross-correlation-based loss function instead of the non-symmetric neural network in BGRL. Please note that not all aforementioned Graph CL approaches require negative samples. Due to the page limits, we provide a more detailed illustration of the related work in App.~\ref{The summarization of related work}. 

In summary, the previous works mainly focused on utilizing different types of augmentations and different degrees of perturbations~\cite{DBLP:conf/nips/YouCSCWS20} to construct sample pairs for contrastive learning while ignoring the order relations among samples. In this paper, we explore the relations under the augmented views that are produced via different degrees of perturbations and improve the pre-training of the graph encoder.

\section{Background}
\label{Background}

\begin{table}[!t]
\centering
\caption{\label{Math_notation} Common mathematical notations}
\renewcommand{\arraystretch}{1.2}
\setlength{\tabcolsep}{1.2mm}{	
\scalebox{1}{
\begin{tabular}{c|l}
\toprule[1.3pt]
Notation       & Explanation    \\\midrule[1pt]
$N$    &  the number of nodes in graph \\
$M$    & the number of augmented views \\
$K$    & the number of negative samples \\
$v_n$  & the $n$-th node in graph\\ 
$\textbf{z}_n$  &  the representation of $v_n$ \\
$\textbf{z}^m_n$ & the representation for the $m$-th view of $v_n$ \\ 
$\hat{\textbf{z}}_n^k$ & the representation of the $k$-th negative sample for $v_n$\\
$s(.,.)$ & the similarity  function \\
$\sigma()_j$ & the $j$-th entry after applying the softmax transformation\\
$\theta$  &  the controlled parameter for the augmentation strategy \\  
$\textbf{g}^c$ & the ground truth score in the coarse ranking\\
$\textbf{s}_n$ & the predicted ranking score in the coarse ranking\\
$\textbf{G}_n$ & the ground truth score for in the fine-grained ranking \\
$\textbf{S}_n$ & the predicted ranking score in the fine-grained ranking\\
$J^c_n$ & the coarse judgment probability \\
$J^f_n$ & the fine-grained judgment probability \\
$J^a_n$ & the C2F judgment probability 
\\ \toprule[1.3pt]
\end{tabular}}}
\vskip-0.15in  \end{table}

In this section, we introduce contrastive learning and reinterpret it from the perspective of learning-to-rank.

\begin{figure*}
    \centering
    \includegraphics[width=0.90\textwidth]{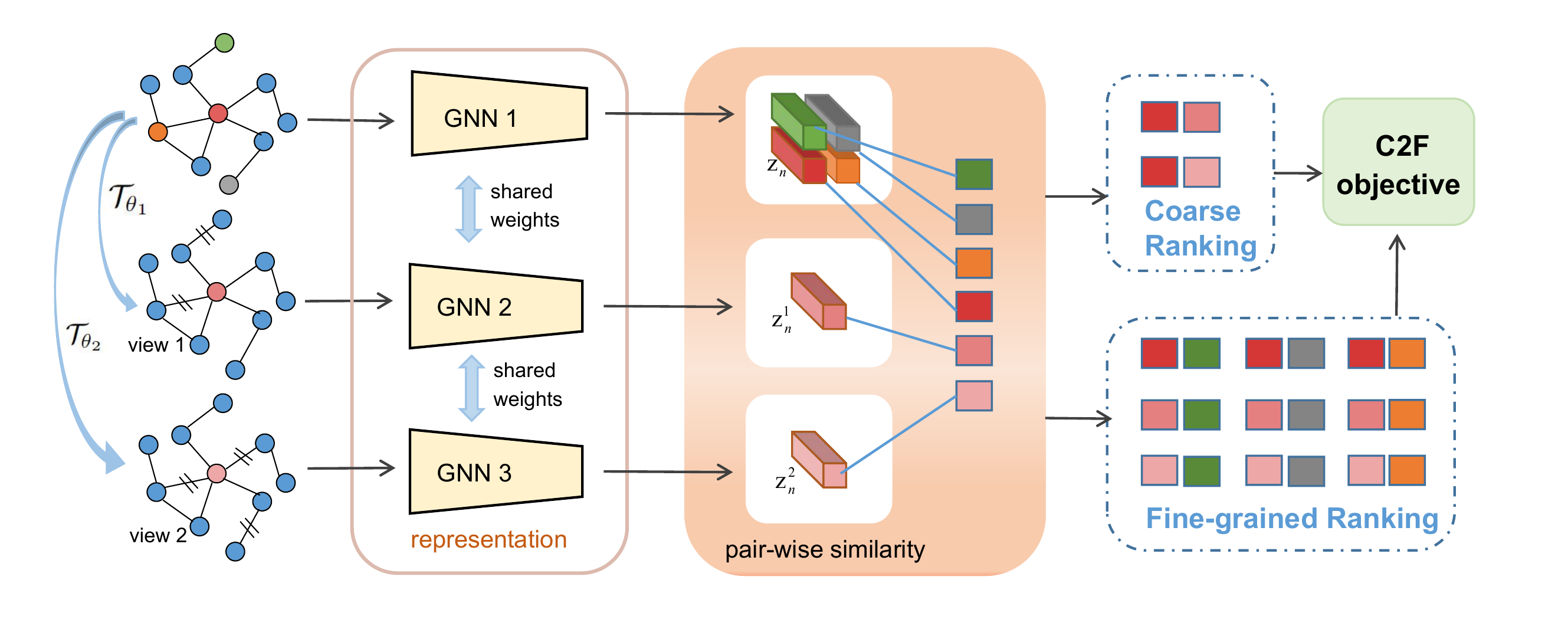}\vskip-0.2in
    \caption{\label{CL_ranking} An illustration of C2F with two augmented views. The two views with different dropping edge ratios and the original graph are sent to the shared GNNs encoder for obtaining node representations. Two augmented nodes $z_n^1$ and $z_n^2$ are positive samples, (i.e., the light red, light pink blocks) are the augmented nodes for the query $z_n$ (i.e., the red block) with the same index in two views, which can generate two positive pairs for computing the coarse ranking loss. While three negative samples (i.e., the green, orange, and grey block) are drawn randomly from the original graph, which can generate 9 pairs for computing the fine-grained ranking loss. Finally, two ranking loss can be unified as the C2F objective function. Best view in color.
    }\vskip-0.2in
\end{figure*}

\subsection{Contrastive Learning on Graphs}

Let $\mathcal{G}=\{ \mathcal{V}, \mathcal{E}, \textbf{X}\}$ represent an undirected attributed graph, where $\mathcal{V}$ denotes the set of $N$ vertices and $\mathcal{E} \subset\mathcal{V} \times \mathcal{V}$ is the set of edge. $\textbf{\textbf{X}}\in R^{N\times d_1}$ denotes the feature matrix of the attributed graph where $\textbf{x}_i\in R^{d_1}$ denotes the feature vector of node $v_i\in \mathcal{V}$. The adjacency matrix $\textbf{A}$ represents topology information of the graph where $\textbf{A}_{i,j}$ is $1$ if an edge exists between node $v_i$ and node $v_j$ and $0$, otherwise. The goal of unsupervised graph representation learning is to train a feature extractor without labels, which can learn node representations $\textbf{Z} \in R^{N\times d_2} $ with matrix $\textbf{A}$ and $\textbf{X}$ as the input. The learned node representations are usually used for downstream tasks such as node classification, link prediction, etc.


Given a query node $v_n$, let $ \textbf{z}^1_n$ and $ \textbf{z}^2_n$ denote its corresponding node representations in two augmented graphs $\mathcal{G}^1$ and $\mathcal{G}^2$ respectively. $\textbf{z}^1_n$ and $\textbf{\textbf{z}}^2_n$ are then referred to as a pair of positive samples. Let $\{\hat{\textbf{z}}^{k}_n\}_{k=1}^K$ denote the node representations of $K$ negative samples randomly selected from the augmented graph $\mathcal{G}^2$. 
Accordingly, the graph contrastive representation learning loss based on InfoNCE ~\cite{DBLP:journals/corr/abs-1807-03748} can be formulated as
\begin{equation}\label{CL_basic}
   \mathcal{L}_{0}= -\frac{1}{N}\sum_{n=1}^N\log \frac{\exp( s( \textbf{z}^1_n, \textbf{z}^2_n))}{\exp( s ( \textbf{z}^1_n, \textbf{z}^2_n)) + \sum_{k=1}^K \exp (s( \textbf{z}^1_n, \hat{\textbf{z}}^{k}_n))}
\end{equation}
where $s(\cdot ,\cdot)$ denotes a parameterized similarity  function~\cite{DBLP:journals/corr/abs-2102-13085}, which maps two vectors to a scalar value. In our work, we build on Memory Bank~\cite{DBLP:conf/cvpr/WuXYL18} that stores pre-computed representations where negative examples are retrieved for the given queries. 

Specifically, CL is developed to maximize the agreement between positive samples from the same instance while minimizing the agreement between samples in a negative pair. Note that the above definition is based on self-supervised node-level representation learning. The global representations of the whole graph or the corrupted graph can be used for constructing the positive and negative pairs in graph-level representation learning (GraphCL~\cite{DBLP:conf/nips/YouCSCWS20}, DGI~\cite{DBLP:conf/iclr/VelickovicFHLBH19}). Next, we will reformulate the graph contrastive learning model (Eq.~\eqref{CL_basic}) as a learning-to-rank process by assigning specific ground truth scores.

\subsection{Learning-to-rank Reinterpretation of Contrastive Learning}\label{CLasL2R}

Learning to rank model, e.g., ListNet~\cite{cao2007learning} aims to learn a ranking function $f$ to order the feature vectors $\textbf{x}_1,\textbf{x}_2\ldots,\textbf{x}_K$ of $K$ objects. ListNet~\cite{cao2007learning} is a general learning-to-rank model, widely applied in document retrieval, collaborative filtering and many other applications. Specifically, ListNet resorts to cross-entropy loss to align the predicted score  $f(\textbf{X})=[f(\textbf{x}_1),f(\textbf{x}_2),\ldots,f(\textbf{x}_K)]$ outputted by the ranking function with the pre-defined ground truth score vector $\textbf{g} = [g_1, g_2, \ldots, g_K]$, where $\textbf{X}=[\textbf{x}_1,\textbf{x}_2,\ldots,\textbf{x}_K]$ is the feature matrix and $g_i$ corresponds to the ground truth score of the $i$-th object. In particular, ListNet adopts the softmax operator $\sigma(\textbf{g})_i = \frac{\exp(g_i)}{\sum_k \exp(g_k)}$ to deliver a normalized probability distribution. The loss function of ListNet can be summarized as follows: 
\begin{equation}\label{listnet}
    D_{CE}(\textbf{g}\|f(\textbf{X}))=-\sum^{K}_{k=1} \sigma(\textbf{g})_k \times \log \sigma(f(\textbf{X}))_k.
\end{equation}
Given two lists of scores, i.e., $\textbf{g}$ and $f(\textbf{X}$), the ListNet loss function (Eq.~\eqref{listnet}) first calculates the probability distributions for ground truth scores and predicted scores, respectively and then measures the distribution divergence via cross-entropy.
\begin{proposition}
InfoNCE~\cite{DBLP:journals/corr/abs-1807-03748} is a specific case of the learning to rank model~\ref{listnet}.
\end{proposition}

\begin{Proof} First, we rewrite the CL loss Eq.~\eqref{CL_basic} as follows:
\begin{align}\label{CL_L2R}
\mathcal{L}_{0}
& =-\frac{1}{N}\sum_{n=1}^N \Big[1 \times \log \frac{\exp (s ( \textbf{z}^1_n,\textbf{z}^2_n))}{\exp (s ( \textbf{z}^1_n,\textbf{z}^2_n))+\sum_{j=1}^K \exp (s( \textbf{z}^1_n,\hat{\textbf{z}}^{j}_n))}\nonumber\\
&  \qquad + \sum_{k=1}^K 0 \times \log \frac{\exp (s ( \textbf{z}^1_n,\hat{\textbf{z}}^{k}_n))}{\exp (s ( \textbf{z}^1_n,\textbf{z}^2_n))+\sum_{j=1}^K \exp (s( \textbf{z}^1_n,\hat{\textbf{z}}^{j}_n))}\Big]\nonumber\\
& \overset{\circled{1}}{=} -\frac{1}{N}\sum_{n=1}^N\Big[\sigma(\textbf{g})_0 \times \log \sigma(\textbf{s}_n)_0+ \sum_{k=1}^{K}\sigma(\textbf{g})_k \times  \log \sigma(\textbf{s}_n)_k\Big]\nonumber\\
& = -\frac{1}{N}\sum_{n=1}^N\Big[\sum_{k=0}^{K} \sigma(\textbf{g})_k \times  \log \sigma(\textbf{s}_n)_k\Big]
\end{align}
where $\circled{1}$ holds by defining the ground truth score $\textbf{g}=[0,-\infty,\ldots,-\infty]$ and the predicted ranking score $\textbf{s}_n = [s ( \textbf{z}^1_n,\textbf{z}^2_n), s ( \textbf{z}^1_n,\hat{\textbf{z}}^{1}_n), \ldots, s ( \textbf{z}^1_n,\hat{\textbf{z}}^{K}_n)] $. Note the index $k$ starts from $0$ to ensure consistency with the definition in contrastive learning Eq.~\eqref{CL_basic}. In detail, since the ground truth score of the positive pair is set to $0$ and the scores of each negative pair are all set to $-\infty$, we have $\sigma(\textbf{g})_0=1$ and $\sigma(\textbf{g})_k=0$ $\forall k =1, \ldots, K$. This is a simple property since both are softmax-based cross-entropies. 
\hfill 
\end{Proof}
\begin{remark}
The aim of the ranking model (Eq.~\eqref{CL_L2R}) is to rank the positive pair $(\textbf{z}^1_n,\textbf{z}^2_n)$ at the top-one, and other negative pairs $(\textbf{z}^1_n,\hat{\textbf{z}}^{K}_n)$ $\forall k =1, \ldots, K$ at the bottom, which is consistent with the goal of contrastive learning: (1) maximizing the agreement between any positive pairs; (2) minimizing the agreement between any negative pairs. 
\end{remark}


However, most of the existing works in CL ignore the relations among augmented views of different magnitudes. Namely, (1) the stronger the perturbation is, the less similar the augmented view and original graph are. (2) CL can not effectively handle a list of augmented views since it is designed for dealing with two of them each time~\cite{DBLP:conf/eccv/TianKI20}. Moreover, the conventional contrastive learning ignore the discriminative information among negative samples. 



\begin{figure*}
    \centering
    \includegraphics[width=0.83\textwidth]{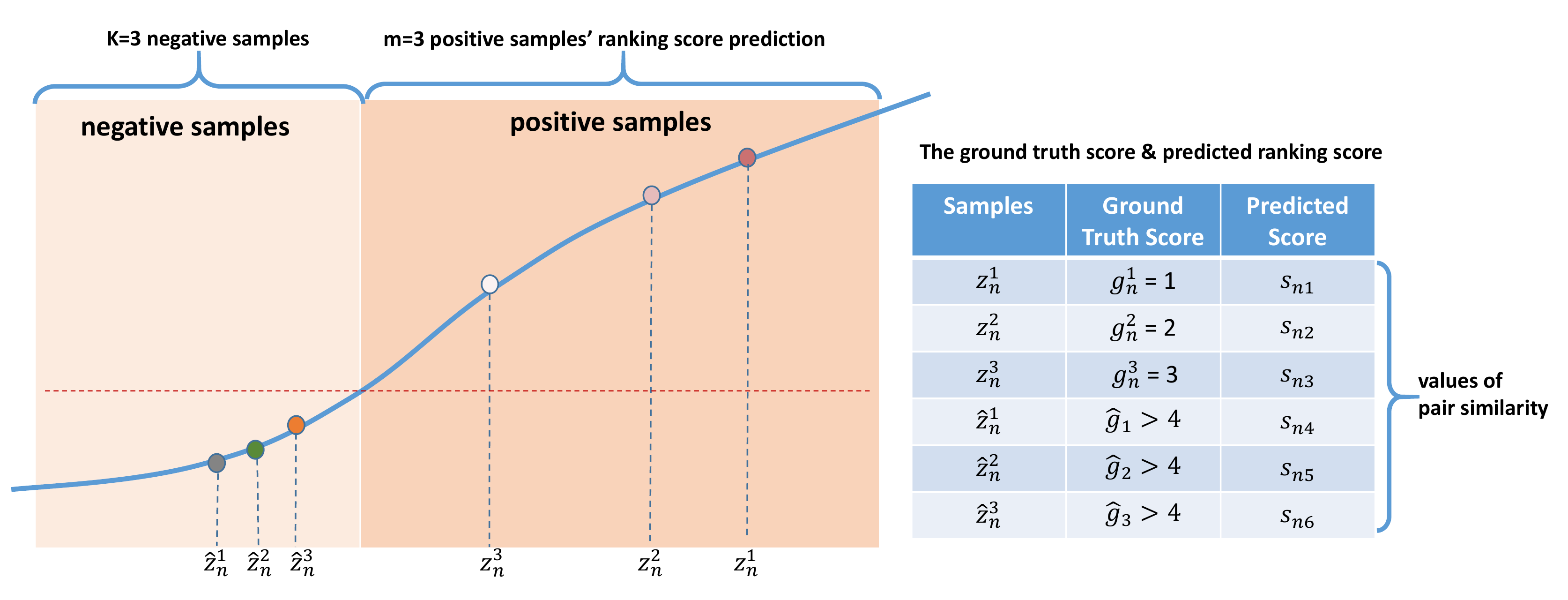}\vskip-0.2in
    \caption{\label{coarse} An illustration of forecasting scheme in the coarse ranking. There are in total three positive samples and three negative samples for the query $\textbf{z}_n$. \textbf{Left:} through optimizing the coarse ranking loss, the optimal ranking function calculates the predicted ranking score of positive and negative samples presented. Three positive samples are ranked at the top where their order is consistent with the order among perturbation degrees of their views. \textbf{Right:} the ground truth score and predicted ranking score are summarized in the right table. The ground truth is from the order in different degrees of perturbations.} \vskip-0.2in
\end{figure*}
\section{Coarse-to-Fine Contrastive learning}
\label{Coarse-to-Fine Contrastive learning}
Taking the above concerns into consideration, we propose a coarse-to-fine contrastive learning (C2F) framework to unify the coarse and fine-grained ranking module. 
The coarse ranking enables better exploitation of a list of augmented views via assigning the order to them according to the degrees of perturbations. Thus the feature extractor is encouraged to learn/encode the order relation among positive views. 
Furthermore, to involve negative samples in ranking, the fine-grained ranking model assigns pseudo-judgments (i.e., self-generated ground truth scores) for negative samples via a self-ranking mechanism. The pre-trained encoder with the supervision of negative samples can prevent the loss of structure information caused by adding perturbation.

Assume $\mathcal{T}_{\theta}(\cdot|\mathcal{G})$ is a graph transformation strategy with a controlled parameter\footnote{In this paper, we utilized only one type of augmentations, i.e., dropping edge, where the drop edge ratio is the controlled parameter.} $\theta$. Given the original graph $\mathcal{G}$, a list of correlated augmented graphs can be constructed via varying $\theta$, namely $\hat{G}=\{{\mathcal{G}}^m |  {\mathcal{G}}^m= \mathcal{T}_{\theta_m}(\cdot|\mathcal{G}), m=1,2,\ldots, M \}$, where $\{\theta_m\}_{m=1}^M$ satisfies the constraint  $\theta_1<\theta_2<\ldots<\theta_M$. 


\subsection{Coarse Ranking}\label{sec:course ranking}



In order to leverage different magnitudes of augmentation, we develop a coarse ranking model by creatively adopting the learning-to-rank method to encode the sorted augmented views. 
According to our observation (see APP.~\ref{insights}), the average node similarity between the original graph and the augmented graph exhibits a decreasing trend with the increase of the perturbation degree. Such prior knowledge is generally applicable and independent of the type of perturbation. Therefore, we propose to incorporate such prior knowledge into CL to enhance the function of the encoder for learning more discriminative node representation.

In particular, given the node representation $\textbf{z}_n$ as the query, let $\textbf{z}_n^1, \textbf{z}_n^2, \ldots, \textbf{z}_n^M$ denote the corresponding node representation of $M$ positive augmented views (ascendingly ordered by their perturbation degrees) and $\hat{\textbf{z}}^{1}_n, \hat{\textbf{z}}^{2}_n, \ldots, \hat{\textbf{z}}^{K}_n$ denote the node representation of $K$ negatives samples. Then, we have the following ranking order in terms of the similarity between the query and each positive/negative sample: 
\begin{equation}\label{coarse_order}
\text{Query: } \textbf{z}_n, \quad \textbf{z}_n^1 > \textbf{z}_n^2 > \ldots> \textbf{z}_n^M \gg \hat{\textbf{z}}^{k}_n,
\end{equation}
where $\hat{\textbf{z}}^{k}_n\in \{\hat{\textbf{z}}^{1}_n,\hat{\textbf{z}}^{2}_n, \ldots, \hat{\textbf{z}}^{K}_n\}$.
Note the degree of perturbation implies no prior knowledge about the discrimination information among negative samples. Thus we model them in the same way as conventional contrastive learning, namely, simply ranking all of them at the bottom. 

Next, we model the ranking order expressed in Eq.~\eqref{coarse_order}. According to the connection of contrastive learning and L2R as discussed in Section~\ref{CLasL2R}, we formulate the coarse ranking model from two aspects: the ground truth (judgment) score and the predicted ranking score, respectively.

\begin{itemize}
    \item \textbf{The ground truth score:} given a list of quantifiable augmentations (Eq.~\eqref{coarse_order}), e.g., a list of augmented views sorted by the degrees of the perturbations on the original graph, the ground truth score $\textbf{g}^c=[g_1, g_2, \ldots, g_M, \hat{g}_1, \hat{g}_2, \ldots, \hat{g}_K]$ for the positive views (the first $M$ entries) as well as the negative samples (the last $K$ entries) should satisfy the following  constraints: 
    \begin{align}
    \begin{cases}g_i> g_j & 1 \le i < j \le M \quad \text{for sorted positive views}\\
     \hat{g}_i = \hat{g}_j & 1 \le i < j \le K\quad \text{for negative samples} \end{cases}  \nonumber
    \end{align}

 Then, we adopt the softmax operator $\sigma(\textbf{g}^c)$ to deliver a normalized ground truth probability. The order among the ground truth $\{g_m\}_{m=1}^M$ is passed from the order of perturbation degree $\{\theta_m\}_{m=1}^M$. For simplicity, we fix the ground truth score for each negative sample to $-\infty$ to ensure they are all ranked at the bottom, i.e., $\hat{g}_i =-\infty$ $\forall i = 1,2,\ldots K$. 
 \item \textbf{The predicted ranking score:} similarly, the similarity function $s(\cdot,\cdot)$ is adopted to calculate the similarity between the query and its positive/negative samples, namely,
\begin{align}
\textbf{s}_{nj} =\begin{cases}
  s(\textbf{z}_n, \textbf{z}_n^j) & 1\le j \le M\\
  s(\textbf{z}_n, \hat{\textbf{z}}_n^{j-M}) & M+1\le j \le M+K\end{cases} 
\end{align}
where $\textbf{s}_n$ is the predicted ranking score of node $v_n$, a vector of dimension $M+K$ for ranking $M$ positive samples and $K$ negative samples simultaneously. 
Then, we can obtain the normalized predicted scores for $M+K$ samples:
\begin{equation}\label{vec_coarse}
 \sigma(\textbf{s}_n)_j = \frac{\exp(\textbf{s}_{nj})}{\sum_{m=1}^{M} \exp(s(\textbf{z}_n, \textbf{z}_n^m)) 
  + \sum_{k=1}^{K}\exp(s(\textbf{z}_n, \hat{\textbf{z}}_n^k))}, 
\end{equation}
which represents the probability of each sample being ranked at the top.
\end{itemize}  

With the ground truth score $\textbf{g}^c$ and the predicted ranking score $\textbf{s}_n$, we define our coarse ranking loss as follows:
\begin{equation}\label{coarse_loss}
 \mathcal{L}_{coarse}=\frac{1}{N}\sum_{n=1}^ND_{CE}(\textbf{g}^c \| \textbf{s}_n).
\end{equation}
where the divergence between two probability distributions can be measured with cross-entropy, optimizing the objective of the coarse ranking loss can obtain the optimal ranking function (i.e., the encoder).

The perturbation degrees $\{\theta_m\}^{M}_{m=1}$ are the available privileged information in training. So the order in perturbation degrees can be regarded as the ground truth score in the coarse ranking model. Fig.~\ref{coarse} illustrates the forecasting scheme of the coarse ranking with three positive views $M=3$ and negative samples $K=3$. 
Due to the proper constraints of ordered ground truths, our coarse ranking loss can capture the order relations among positive views and guide the encoder to learn discriminative information in a self-supervised ranking manner. However, the coarse ranking still ignores the fact that the augmentation strategy will distort graph structure to a greater extent when the perturbation degree increases.

\subsection{Fine-grained Ranking}
The augmentation strategy will inevitably corrupt the graph's topological structure when randomly dropping edges with a certain percentage.
According to our observation (see in APP.~\ref{insights}), the similarity between nodes will decrease within the same augmented view as more and more edges are deleted. The loss of structure information will enlarge the distance between nodes, which goes against the intention of enhancing node smoothness in graph neural networks. 
In order to tackle the challenge, we further propose a fine-grained ranking model that supervises the fitting of augmented view via incorporating negative samples into sorting. 

\begin{figure}[!t]
    \centering
    \includegraphics[width=0.51\textwidth]{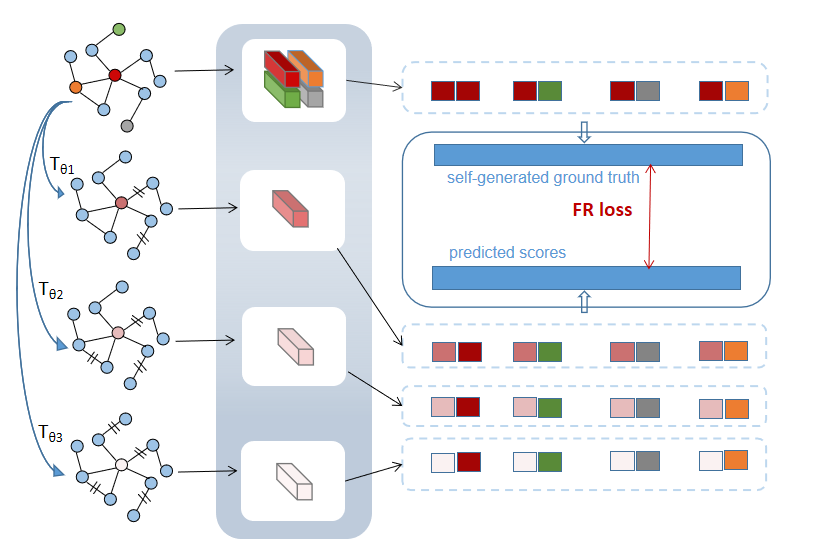}\vskip-0.15in
    \caption{ \label{fine}Illustration of pair generation in the fine-grained ranking loss (FR loss). There are in total 3 augmented views and 3 negative samples (the green, orange, and grey block). The positive (the red block) and negative samples are from the original graph, which can compose 4 pairs with the query (the red block). Their similarities are regarded as the self-generated ground truth scores of other pairs in augmented views. Best view in color.}\vskip-0.2in
\end{figure}
We devise a self-supervised ranking mechanism to assign self-generated ranking supervisions for negative samples of query $\textbf{z}_n$. First we randomly draw $K$ negative samples from the original graph to construct $K$ negative pairs $\{(\textbf{z}^m_n, \hat{\textbf{z}}_n^k)\}_{k=1}^K$ for each positive sample $\textbf{z}^m_n$. Those negative samples are shared to all positive samples $\{z_n\}^M_{m=1}$. The pair generation is shown in Fig.~\ref{fine}. We propose to use $s(\textbf{z}_n,\hat{\textbf{z}}_n^k)$, the score of the corresponding negative pair in the original graph as the ground truth score of the negative pair $\{s(\textbf{z}_n^m,\hat{\textbf{z}}_n^k)\}^M_{m=1}$ in $M$ augmented view. 
Similar to the definition in Sec.\ref{sec:course ranking}, 
we let $\textbf{g}_n=[g_0, \hat{g}_1, \hat{g}_2, \ldots, 
\hat{g}_K]$ denote the ground truth score in the fine-grained ranking module with its first entry and the other $K$ entries indicating the ground truth score of the positive pair and $K$ negative pairs, respectively, written as:
\begin{align}\label{vec_rank}
    \textbf{g}_{nk} =\begin{cases} 
     s(\textbf{z}_n,\textbf{z}_n) & k = 0\\
     s(\textbf{z}_n,\hat{\textbf{z}}_n^k) & 1\le k \le K \quad 
     \end{cases}   
    \end{align}
Meanwhile, for each augmented view, the predicted ranking score vector is denoted as:
 \begin{align}\label{mat_score}
     \textbf{s}_{nk}^m =\begin{cases}
     s( \textbf{z}_n^m, \textbf{z}_n) & k=0\\
     s(\textbf{z}_n^m, \hat{\textbf{z}}_n^k) & 1\le k \le K
     \end{cases} 
 \end{align}
Note there are $M$ augmented views with each one containing $K+1$ entries. To obtain a normalized top-one ranking probability, we first formulate $\textbf{S}_n$ as a matrix with $\textbf{s}^m_{nj}$ denoting the entry of $\textbf{S}_n$ at the $m$-th row and the $j$-th column, $\forall m =1,2,\ldots, M, \ j =0,1,\ldots, K$.  Then, we introduce a normalization formulation of matrix version for the predicted scores $S$, namely 
\begin{equation}\label{mat_pred}
     \sigma(\textbf{S}_n)^m_j = \frac{\exp(\textbf{s}^m_{nj})}{\sum_{m=1}^{M} \big(\exp(s(\textbf{z}_n, \textbf{z}_n^m)) 
     + \sum_{k=1}^{K}\exp(s(\textbf{z}^m_n, \hat{\textbf{z}}^k_n)) \big)}. 
\end{equation}
To adapt the ranking supervision $ \textbf{g}_{n}$ (Eq.~\eqref{vec_rank}) for the predicted score $\textbf{S}_n$, we introduce the augmented ranking supervision $ \textbf{G}_{n}$.  and its normalized top-one ranking probability as follows
\begin{subequations}
\begin{align}
\textbf{G}_n &= \mathbf{1} \ast \textbf{g}_n^T, \label{mat_rank}\\
\sigma(\textbf{G}_n)_j^m &= \sigma(\mathbf{1} \ast \textbf{g}_n^T)_j^m = \frac{1}{M}\big(\mathbf{1} \ast \sigma(\textbf{g}_n)^T\big)_j^m, \label{mat_prob}
\end{align}
\end{subequations}
where $\mathbf{1}$ is a $M \times 1$ vector with all elements equal to one and $*$ indicates the matrix multiplication.
Our fine-grained ranking loss is then defined as 
\begin{equation}\label{fine_loss}
\begin{split}
 \mathcal{L}_{fine}&=\frac{1}{N}\sum_{n=1}^ND_{CE}(\textbf{G}_n\| \textbf{S}_n) \\
 D_{CE}(\textbf{G}_n\| \textbf{S}_n)&=-\sum_{m=1}^M\sum_{j=0}^K \sigma(\textbf{S}_n)^m_j \times \log \sigma(\textbf{G}_n)_j^m 
\end{split}
\end{equation}
where the cross-entropy loss between two probability matrices is calculated element-wise. To summarize, in the fine-grained ranking module, we provide pseudo-judgments for negative samples to supervise the fitting of augmented views for avoiding the structure information loss.

\subsection{Unifying Both as the Coarse-to-Fine Ranking}
In this section, we focus on building a general learning-to-rank model for contrastive graph representation learning. 

To make full use of the strengths of the two aforementioned modules, we simply combine the two objectives as follows 
\begin{equation}\label{final loss}
    \mathcal{L}_{CF}=\mathcal{L}_{coarse}+\lambda\mathcal{L}_{fine},
\end{equation}
where $\lambda > 0$ is the weight factor that balances these two ranking loss.

However, the unified objective (Eq.~\eqref{final loss}) no longer fits the definition of ListNet (Eq.~\eqref{listnet}). Therefore, we revisit the definition of both coarse and fine-grained objectives to deliver a well-defined learning-to-rank formulation. 

First of all, comparing the normalized prediction score for coarse (Eq.~\eqref{vec_coarse}) and fine-grained (Eq.~\eqref{mat_pred}) models, the difference lies in the second term of their denominators, namely $\sum_{k=1}^{K}\exp(s(\textbf{z}_n, \hat{\textbf{z}}_n^k))$  vs. $\sum_{m=1}^M \sum_{k=1}^{K}\exp(s(\textbf{z}^m_n, \hat{\textbf{z}}^k_n))$, that hinders us from combining similar terms. 

To achieve consistency between the two denominators, we propose to use the denominator in Eq.~\eqref{mat_pred} as the standardized denominator for both coarse and fine-grained normalized prediction scores while keeping their numerators unchanged, respectively. 

\begin{remark} [Justification for the choice of the new denominator]
  In the following, we justify the above choice from three aspects: (1) the modification is only on the choice of negative pairs. It will not affect the definition of coarse ranking loss, since its ground truth score for each negative pair is fixed to zero. 
  (2) The new denominator contributes to a better estimation of the mutual information. According to the definition of InfoNCE (Eq.~\eqref{CL_basic})~\cite{DBLP:journals/corr/abs-1807-03748}, the larger the number of negative pairs ($M*(K+1)$ Vs. $K+1$), the better approximation of the contrastive learning loss for mutual information. (3) The similarity between the augmented views and negative samples (i.e.,$s(\textbf{z}_n^m, \hat{\textbf{z}}_n^k)$) are assumed to be comparable to the similarity between the original node and negative samples (i.e.,$s(\textbf{z}_n,\hat{\textbf{z}}_n^k)$, $\forall k = 1,2, \ldots, K$) according to the definition of fine-grained ranking loss. The new denominator has the same scale as the previous ones when reaching to the optima. 
\end{remark}
Therefore, we introduce the modified loss as follows:
\begin{equation}
\begin{aligned}\label{overall_loss}
    \mathcal{L}_{CF} &=\frac{1}{1+\lambda}\left(\hat{\mathcal{L}}_{coarse}+\lambda\mathcal{L}_{fine}\right) \\
    & = \alpha\hat{\mathcal{L}}_{coarse}+ (1-\alpha)\mathcal{L}_{fine},
\end{aligned}
\end{equation}
where $\hat{\mathcal{L}}_{coarse}$ denotes the modified loss for the coarse model, which replaces the denominator in Eq.~\eqref{vec_coarse}) with that in Eq.~\eqref{mat_pred}. The coefficient $\frac{1}{1+\lambda}$ is introduced to ensure the judgment probability matrix is normalized to one. And $\alpha = \frac{1}{1+\lambda}$. The algorithm of our coarse-to-fine contrastive learning is described in Algorithm 1. 

In more detail, let $J^{c}_n$, $J^{f}_n$ and $J^{a}_n$ denote the coarse, fine-grained and overall judgment probability matrix in $\hat{\mathcal{L}}_{coarse}$, $\mathcal{L}_{fine}$ and $\mathcal{L}_{CF}$ in Eq.~\eqref{overall_loss}, respectively. Then we have 
\begin{subequations}
 \begin{align}
 J^c_n & = \sigma(\textbf{g}^c_{1:M} \ast \textbf{a}^T) = \sigma(\textbf{g}^c_{1:M}) \ast \textbf{e}^T, \label{J_c}\\
 J^f_n &= \sigma(\mathbf{1} \ast \textbf{g}_n^T) = \frac{1}{M}\ast \mathbf{1} \ast \sigma(\textbf{g}_n)^T, \label{J_f}\\
  J^a_n &= \alpha J^c_n + (1-\alpha)J^f_n,\label{J_a}
  \end{align} 
\end{subequations}
where $\textbf{g}^c_{1:M}$ denotes the first $M$ entries of $\textbf{g}^c$, which contains the ground truth score for $M$ positive views. $\textbf{a}$ is a $(K+1) \times 1$ vector with the first entry equaling $1$ and the rest entries equaling $-\infty$. $\textbf{e}$ is a $(K+1) \times 1$ vector with the first entry equaling $1$ and the rest entries equaling $0$. We specifically rewrite the matrix $J^a_n$ in section ~\ref{discuss}. Our proposed Coarse-to-Fine ranking can satisfy the ListNet ranking by proving that $J_n^a$ is a normalized judgment probability matrix in APP.~\ref{The analyze of Coarse-to-Fine ranking}.

\begin{algorithm}[!t]
\caption{\label{DECO_Algo}coarse-to-fine contrastive learning on graphs.}
\begin{algorithmic}[1]
\STATE \textbf{Input}: the graph $\mathcal{G}$, the ranking list of augmentation parameter $\Theta=\{\theta^m\}_{m=1}^M$, the augmentation strategy $\mathcal{T}_{\theta}$, similarity  function $s(\cdot,\cdot)$, encoder function $f$, hyperparameter $\alpha$.\\
	\WHILE{Not Converge}
	\STATE $\textbf{Z}=h(\mathcal{G})$ {\color{gray}\# apply encoder}\\
	\STATE $\mathcal{G}^m = \mathcal{T}_{\theta^m}(\mathcal{G})$, $m = 1,2,\ldots,M. ${\color{gray}\# create M views}\\
	\STATE $Z^m = h(\mathcal{G}^m)$, $m = 1,2,\ldots,M.\quad ${\color{gray}\# apply encoder}
	\FOR{$n=1,2,\ldots,N$}
	\STATE $\hat{\textbf{z}}^k_n \sim \textbf{Z},k=1,\dots,K$ {\color{gray}\# choose negative samples}\\
	{\color{gray}\# predicte score for C2F rank}\\
	\STATE {calculate
	$\textbf{S}_n$ following Eq.~\eqref{mat_score}}\\
	 {\color{gray}\# ground truth score for C2F rank}\\
	\STATE {calculate $J^c_n$ following Eq.~\eqref{J_c}}\\
	\STATE {calculate $J^f_n$ following Eq.~\eqref{J_f}}\\
	\STATE {calculate $J^a_n$ following Eq.~\eqref{J_a}}\\
	\ENDFOR
	\STATE{calculate the C2F ranking loss following Eq.~\eqref{overall_loss}}
	\STATE{update encoder $f$ to minimize $\mathcal{L}_{CF}$}
	\ENDWHILE
\STATE \textbf{return} the encoder $f$ and embedding matrix $\textbf{Z}$.
\end{algorithmic}
\end{algorithm}

\section{Discussion}\label{discuss}

In our coarse-to-fine contrastive learning model, the probability matrix of judgment can be defined detailedly as follows,
\begin{align}\label{judgment}
J^a_n &=\alpha J^c_n + (1-\alpha)J^f_n =\alpha\begin{pmatrix}
\sigma(\textbf{g}_{1:M}^c)_{1} &  0& \ldots  &  0\\
\vdots             & \vdots &   \ddots  & \vdots   \\
\sigma(\textbf{g}_{1:M}^c)_{M}  &  0& \ldots  &  0\\
\end{pmatrix} \nonumber\\
&\quad
+ \frac{(1-\alpha)}{M}\begin{pmatrix}
\sigma(\textbf{g}_n)_{0} &  \sigma(\textbf{g}_n)_{1}   & \ldots  &\sigma(\textbf{g}_n)_{K}\\
\vdots  & \vdots &   \ddots  & \vdots   \\
\sigma(\textbf{g}_n)_{0}    & \sigma(\textbf{g}_n)_{1}  & \ldots  & \sigma(\textbf{g}_n)_{K}\\
\end{pmatrix} 
\end{align}

It plays an important role in guiding the pre-training process of the encoder. In the following, we discuss the meaning of each row and column in the judgment probability matrix $J_n^a$.
\begin{itemize}
    \item \textbf{Each row of $J^a_n$ can be viewed as a ranking list of judgment probability of the $m$-th view}. 
    
    (1) Each row corresponds to a ranking list of sample pair similarities, $[ s( \textbf{z}_n^m, \textbf{z}_n),s(\textbf{z}_n^m, \hat{\textbf{z}}_n^1),\dots,
     s(\textbf{z}_n^m, \hat{\textbf{z}}_n^K)]$.
      
    (2) The judgment probability of positive pairs, $\alpha\sigma(\textbf{g}_{1:M}^c)_{m} + \frac{(1-\alpha)}{M}\sigma(\textbf{g}_n)_{0}$, is larger than that of negative pairs $\frac{1-\alpha}{M}\sigma(\textbf{g}_n)_k$, which implicates that C2F maximizes the agreement between positive samples and minimize the agreement between negative samples.
     
    \item \textbf{Each column of $J^a_n$ can be viewed as a ranking list of judgment probability for each sample (i.e., $\textbf{z}_n$ and $\{\hat{\textbf{z}}_n^k\}_{k=1}^K$)}. 

    (1) The first column corresponds to a ranking list of positive pair similarities, $[s( \textbf{z}_n^1, \textbf{z}_n),s( \textbf{z}_n^2, \textbf{z}_n),\dots,s(\textbf{z}_n^M, \textbf{z}_n)]$ where the order relation as priors can be incorporated into the coarse ranking model. Furthermore, other column $k=1,\dots,K$ corresponds to a list of negative pair similarities, $[s(\textbf{z}_n^1, \hat{\textbf{z}}_n^k), s(\textbf{z}_n^2, \hat{\textbf{z}}_n^k), \dots,s(\textbf{z}_n^M, \hat{\textbf{z}}_n^k)]$, respectively.
    
    (2) The first column represents a ranking list in decreasing order of the judgment probability where the judgments satisfy the following relation:
    \begin{equation}
    \frac{\alpha \sigma(\textbf{g}_{1:M}^c)_{i}}{\alpha \sigma(\textbf{g}_{1:M}^c)_{j}}> \frac{\alpha \sigma(\textbf{g}_{1:M}^c)_{i}+\frac{(1-\alpha)}{M}\sigma(\textbf{g}_n)_0}{\alpha \sigma(\textbf{g}_{1:M}^c)_{j}+\frac{(1-\alpha)}{M}\sigma(\textbf{g}_n)_0}>1,\label{colimpli}
    \end{equation}
    where $1\le i<j\le M$. Eq.\eqref{colimpli} implicates that C2F is able to capture the relation among ordered views. Moreover, for the $k$-th ($k\neq0$) column in $J^a_n$, the judgment probability is equal to $\frac{(1-\alpha)}{M}\sigma(\textbf{g}_n)_{k}$. Thus the pre-training process of the encoder is supervised by allowing negative samples to involve in ranking, which can avoid the loss of structure information.

\end{itemize}



\section{Experiment}
\label{Experiment}
In this section, we conduct experiments to verify the superiority of our Coarse-to-Fine contrastive learning (C2F). We pretrain the GNNs encoder on six graph datasets and evaluate its performance on the node classification task. As C2F is a general framework and independent of data type, we also evaluate the quality of the pre-trained ResNet-50 encoder on an image dataset. The detailed experiments of C2F on images are shown in APP.~\ref{image1},APP.~\ref{image2} and APP.~\ref{B}.

\subsection{Datasets and Baselines} 
\textbf{Datasets.} We use six widely used benchmark datasets in our experiments: a citation network (i.e., Pubmed), a web page dataset (i.e., Facebook)~\cite{DBLP:journals/compnet/RozemberczkiAS21}, two segments of the Amazon
co-purchase graph (i.e., Amazon-Com, Amazon-Photo)~\cite{DBLP:conf/sigir/McAuleyTSH15} and two coauthor datasets (i.e., Coauthor-CS and Coauthor-Phy)~\cite{DBLP:journals/corr/abs-1811-05868}. The first five datasets are for the node classification task in the transductive setting, but the last dataset is for the inductive setting (i.e., the test nodes are not seen during training). We follow CGNN~\cite{DBLP:journals/corr/abs-2008-11416} to split six benchmark datasets, which has been widely adopted in semi-supervised learning on graph. We provide detailed discussions about the dataset split in APP.~\ref{split}.  
We show the data statistics in~Table~\ref{Rank_4}. 
\begin{table}[htbp!]
\centering
\caption{\label{Rank_4} The statistics of dataset.}
\setlength{\tabcolsep}{1.2mm}
\scalebox{0.90}{
\begin{tabular}{ccccccc}\toprule[1.5pt]
Dataset & \#Nodes & \#Edges & Density & \#Features & \#Classes & Train/Val/Test \\\midrule[1.0pt]
Pubmed  & 19,717 & 44,338  & 0.01\% & 500 & 3 & 60/500/1,000  \\
\hline
Facebook & 22,470 & 170,823  & 0.03\% & 4,714 & 4 & 80/120/rest \\ 
\hline
Amazon-Com  &  13,752 & 245,861  & 0.13\% & 767 & 10 & 200/300/rest  \\
\hline
Amazon-Pho  &  7,650 & 119,081  & 0.20\% & 745 & 8 & 160/240/rest  \\
\hline
Coauthor-CS & 18,333 & 81,894 & 0.02\% & 6,805 & 6 & 300/450/rest \\
\hline
Coauthor-Phy & 34,493 & 247,962 & 0.02\% & 8,415 & 5 & 20,000/5,000/rest\\
\bottomrule[1.5pt]
\end{tabular}}
\vskip-0.05in  \end{table}

\textbf{Baselines.} We adopt dropping edge as the augmentation strategy of C2F, which is referred to as C2F. C2F using feature masking to generate augmented views is referred to as C2F w/ FM. We compare our methods with self-supervised graph representation learning methods based on contrastive learning (i.e., DGI~\cite{DBLP:conf/iclr/VelickovicFHLBH19}, MVGRL~\cite{DBLP:conf/icml/HassaniA20}, GCC~\cite{DBLP:conf/kdd/QiuCDZYDWT20}, CGNN~\cite{DBLP:journals/corr/abs-2008-11416}, GCA~\cite{DBLP:conf/www/0001XYLWW21}, GRACE~\cite{zhu2020deep}, BGRL~\cite{thakoor2021bootstrapped}, G-BT~\cite{DBLP:conf/icml/ZbontarJMLD21}). We also choose several common unsupervised graph representation learning models as baselines (i.e., DeepWalk~\cite{DBLP:conf/kdd/PerozziAS14}, Node2Vec~\cite{DBLP:conf/kdd/GroverL16}). To demonstrate the potential of C2F, we compare C2F with the popular supervised GNN models (i.e., GCN~\cite{DBLP:conf/iclr/KipfW17}, GAT~\cite{DBLP:conf/iclr/VelickovicCCRLB18}, GraphSage~\cite{DBLP:conf/nips/HamiltonYL17}). Those baselines can be summarized as follows:
\begin{itemize}
\item DGI~\cite{DBLP:conf/iclr/VelickovicFHLBH19}: constructs a negative view by shuffling the attribute matrix and minimizes the Jensen-Shannon divergence between the joint and the product of marginals.~\cite{DBLP:conf/iclr/HjelmFLGBTB19}. 
\item MVGRL~\cite{DBLP:conf/icml/HassaniA20}: utilizes graph diffusion to create positive views and maximizes mutual information between the node embeddings and graph embedding in the positive views.
\item GCC~\cite{DBLP:conf/kdd/QiuCDZYDWT20}): generated sub-graph of each node via random walk as instance and utilizes contrastive learning to achieve instance discrimination.  
\item CGNN~\cite{DBLP:journals/corr/abs-2008-11416}: replaces the softmax function in contrastive learning with noise contrastive estimation for learning node representations. 

\item GCA~\cite{DBLP:conf/www/0001XYLWW21}: designs adaptive augmentation on the graph topology and node attributes to incorporate priors of topological and semantic aspects of the graph.
\item GRACE~\cite{zhu2020deep}: corrupts the graph structure and node attributes to generate two views and uses other nodes from two augmented views as negative samples in the InfoNCE objective function.
\item BGRL~\cite{thakoor2021bootstrapped}: extends BYOL~\cite{DBLP:conf/nips/GrillSATRBDPGAP20}, the outstanding CL method on images, to graphs. It avoids the use of negative pairs and instead utilizes various tricks, like stop gradient, predictor network, and momentum encoders to improve performance.
\item G-BT~\cite{DBLP:journals/corr/abs-2106-02466}: extending the CL method on images, Barlow Twins~\cite{DBLP:conf/icml/ZbontarJMLD21}, to graphs, uses a cross-correlation-based loss instead of the non-symmetric neural network of BGRL.
\item DeepWalk~\cite{DBLP:conf/kdd/PerozziAS14}, Node2Vec~\cite{DBLP:conf/kdd/GroverL16}: are widely used conventional unsupervised methods of graph representation learning. 
\item GCN~\cite{DBLP:conf/iclr/KipfW17}, GAT~\cite{DBLP:conf/iclr/VelickovicCCRLB18}, and GraphSage~\cite{DBLP:conf/nips/HamiltonYL17}: are widely used supervised graph representation learning.
\end{itemize}

We conduct experiments of C2F on graphs following CGNN, where we employ GAT~\cite{DBLP:conf/iclr/VelickovicCCRLB18} as the GNN-based backbone.
Besides, we also conduct experiments of C2F on CIFIR-10~\cite{Krizhevsky2009LearningML} (an image dataset) following SimCLR~\cite{DBLP:conf/icml/ChenK0H20} and employs ResNet-50 as backbone.

\begin{table*}[!t]
\centering
\caption{\label{Rank_6} Comparison of C2F with supervised/unsupervised baselines on node classification accuracy (\%). ``FM" denotes feature masking.} \vskip-0.05in
\setlength{\tabcolsep}{1.2mm}
\scalebox{1.0}{
\begin{tabular}{c|ccccccc}\toprule[1.5pt]
& Method & Pubmed & Facebook & Amazon-Com & Amazon-Pho & Coauthor-CS & Coauthor-Phy \\\midrule[0.8pt]
 & DeepWalk  & 65.59 & 63.04 & 76.93 & 81.50 & 77.81 & 91.17 \\

& Node2Vec   & 70.34 & 69.69 &  75.49 & 82.21 & 79.93 & 91.43   \\

& DGI  & 79.24 &  69.53 &  71.41 & 79.34 &  91.41 & 93.26 \\

Unsupervised & MVGRL  & 80.10 &  67.24 &  67.15 & 79.54 & 90.73 & 91.49  \\

& GCC  & 80.60 & 70.36 & 74.18 & 83.60 & \bf{91.76} & 93.97   \\

 & CGNN & 80.93 & 78.39 &  75.11 & 89.84 & 90.14 & 92.34  \\

 & GCA & 80.02 & 71.52 & 77.30 & 85.50 & 90.97 & 73.09\\
 & GRACE~\cite{zhu2020deep} & 78.10 & 65.81 & 69.32 & 66.48 & 90.20 & 72.90 \\
 & BGRL~\cite{thakoor2021bootstrapped} & 71.01 & 62.42 &  \bf{81.72} & 86.02 & 89.31 & 75.85   \\
 & G-BT~\cite{DBLP:journals/corr/abs-2106-02466}  & 80.04  & 64.14 &  75.17 & 84.49 & 90.38 & 74.05 \\
& \bf{C2F w/ FM} & 80.05 & 79.56 & 76.05 & 87.60 & 90.40 & 93.72  \\
& \bf{C2F}  & \bf{81.10} & \bf{79.92} & 77.79 & \bf{89.88} & 91.07 &  \bf{94.09}  \\

\hline
& GCN & 79.20 &  66.37 &  81.18 &  85.82 & 92.01 & 93.35 \\

Supervised & GAT & 78.71 &  72.24 & 81.85  &  87.46 & 91.23 &  95.89 \\

& GraphSage & 79.02 & 69.62 & 82.10 & 87.60 & 92.60 & 95.28  \\

& GAT(DropEdge) & 78.90 & 71.57 & 82.20 & 87.59 &  91.32 & 93.80  \\
\bottomrule[1.0pt]
\end{tabular}} \vskip-0.1in
\end{table*}

\subsection{Experimental Setting}
We implement our method with Pytorch on a 24GB GeForce RTX 3090 GPU. We follow the official codes of CGNN~\cite{DBLP:journals/corr/abs-2008-11416} and keep the hyperparameters unchanged. Specifically, for the GAT layer in C2F, we follow~\cite{DBLP:conf/iclr/VelickovicCCRLB18} to set the number of attention heads to be 8, the number of units in each attention head to be 8, and the number of GAT layers to be 2. In the pre-training process, all models are initialized from uniform distribution and trained to minimize the C2F loss using Adam optimizer with a learning rate of 0.001 for 5,000 iterations. The similarity function for C2F is formalized as $s(\textbf{z}_1,\textbf{z}_2)=\textbf{z}_1^T\textbf{z}_2/\tau$, where the temperature $\tau$ is set to 0.1. The negative sampling size $K$ is empirically set to 1024. In C2F, the dropping-edge ratios of two augmented views are set to $\{0.5,0.8\}$, the judgments are set to $\{1.0,0.7\}$, and the balance parameter is set to $0.8$.

Besides the dropping-edge augmentation in C2F, we also adopt feature masking as the alternative augmentation strategy. We follow the feature masking of G-BT, which generates a mask of size $d_1$ (the dimension of node attributes) sampled from the Bernoulli distribution $\mathcal{B}(1-p_X)$. The features of the same dimension are masked for each node, where $p_X$ is the probability of feature masking. In \textit{C2F w/ FM}, two probabilities, $p_{X}^1$ and $p_{X}^2$ of feature masking are set to $\{0.8, 0.5\}$ to generate two ordered views. We set the probabilities of dropping edges to be $\{0, 0\}$, judgment scores to be $\{1, 0.7\}$ and the balance parameter to be $0.8$. Other hyper-parameters are provided in APP.~\ref{Hyper-parameter}.


\subsection{Comparison with Baselines on Graphs}
The experiment results on six real graph-structured datasets are provided in Table~\ref{Rank_6}. We perform C2F based on two augmented views with dropping-edge ratios $\{0.5, 0.8\}$, where the balance parameter $\alpha$ is set to $0.8$, and the judgments are set to $\{1, 0.7\}$ in Eq.~\ref{coarse_loss}. Besides accuracy, we provide the results of various algorithms on additional metrics, i.e., F1, AUC, and Recall score in APP.~\ref{metrics}.

In general, our C2F achieves better performance compared to most of graph CL methods (i.e., DGI, MVGRL, GCC, CGNN, GCA, GRACE and G-BT) on six benchmark datasets and outperforms traditional unsupervised graph methods (i.e., DeepWalk, Node2Vec) significantly. Compared to approaches without negative samples (i.e. BGRL, G-BT), our C2F can achieve better performance on five benchmark datasets. BGRL achieves better performance on the Amazon-Com dataset. We believe this is because Amazon-Com is a graph with a relatively high density, leading to relatively smoothed node representations and false negative samples to deteriorate C2F performance, while BGRL does not rely on negative samples.

Additionally, both C2F and C2F w/ FM perform better than approaches that use negative samples (i.e., GCC, CGNN, MVGRL, GRACE). C2F performs better than C2F w/ FM. We believe it is because the topology of a graph contains more information than node attributes in the context of graph representation learning. C2F achieves better results on all datasets than CGNN. We assume that the benefit stems from the fact that, C2F encourages the graph encoder to capture the prior of the order relationship among positive and negative samples, which can enhance discriminative node representation learning, while the previous CL methods are limited to extracting these priors during pre-training.

C2F is not superior to GCC~\cite{DBLP:conf/kdd/QiuCDZYDWT20} on the Coauthor-CS dataset due to the use of different augmentation strategies. 
Different from C2F, GCC utilizes \textit{the random walk with restart} ~\cite{DBLP:conf/icdm/TongFP06} as its augmentation strategy to generate two sub-graphs (augmented views) for the vertex $v$, which can retain more edges especially benefiting representation learning on sparse graphs ~\cite{DBLP:conf/iclr/RobinsonCSJ21}, e.g., Coauthor-CS. Instead of exploring multi-type augmentation strategies to boost performance in the baselines, e.g., GRACE, GCA, and MVGRL, our proposed models focus on how to incorporate prior information from single-type augmentation by properly sorting the positive and negative samples in the proposed ranking loss for contrastive learning. 

We also observe that C2F w/ FM is not superior to CGNN on the Pubmed and Amazon-Photo datasets. We believe it is because they apply different augmentations on the two datasets with low-dimensional node features\footnote{Amazon-Com is also one dataset with low-dimensional node features, but it has twice nodes on average in each class than Amazon-Pho, which means Amazon-Com has more feature information to benefit C2f w/ FM for training the classifier in the downstream node classification task.}, where the topology structure in graphs contains more useful information than node features for node representation learning. So feature masking adopted by C2F w/ FM provides less prior information to help the encoder with training than dropping edge adopted by CGNN.

Moreover, C2F appears competitive with the supervised models and even outperforms them in Pubmed, Facebook, and Amazon-Pho, verifying the potential of our proposed coarse-to-ﬁne ranking strategy for CL on the node classiﬁcation task. For example, our method improves by 1.47\% on average, compared to GCN. It illustrates that our C2F can leverage more useful prior information in unsupervised learning without the label information. C2F shows comparative performance compared to GAT with the dropping-edge strategy. 

\subsection{Ablation Study on Graphs}
In this section, we focus on the performance of each component in C2F, including the coarse and fine-grained ranking module. When the balance parameter $\alpha = 1$ is set to $1$ in C2F, the coarse ranking is defined. When $\alpha \in [0,1)$ and the judgments of two views are set to $1$, the fine-grained ranking is defined. We conduct experiments of C2F with two views and investigate the effectiveness of each component.
The experimental results are represented in Table~\ref{Rank_8}. The first row of the table denotes the results of vanilla contrastive learning which don't contain any ranking models via setting $\alpha=1$, the judgments of two views to $1$ and the preserve edge ratios of two views to $0.8$. The results in the last row of Table~\ref{Rank_8} consider the whole C2F ranking model with dropping edge.

Specifically, compared to the vanilla contrastive learning in the first row, the coarse ranking gets better performance, which verifies the effectiveness of coarse ranking. The improvement of the fine-grained ranking on average reaches 1.423\% compared to vanilla contrastive learning, which illustrates the effectiveness of fine-grained ranking. Moreover, it is apparently shown that C2F achieves the top performance on all datasets. And we conclude that both components are important for the model to capture more useful prior information. 

\begin{table*}[htbp!]
\centering
\caption{\label{Rank_8} The classification accuracy(\%) of different components of C2F.}\vskip-0.1in
\setlength{\tabcolsep}{1.2mm}
\scalebox{0.9}{
\begin{tabular}{c|cccccc|cc}\toprule[1.5pt]
Dataset & Pubmed & Facebook & Amazon-Com & Amazon-Pho & Coauthor-CS & Coauthor-Phy &  Coarse & Fine-grained  \\\midrule[1.0pt]
\multirow{4}{*}{Acc(\%)} & 79.02 & 76.95 & 75.72 & 87.95 & 89.00 &  91.68 & $\cdot$ & $\cdot$ \\
 & 79.70  & 78.45 & 76.38 & 89.39 & 90.88 & 93.85  & $\checkmark$ & $\cdot$    \\
 & 79.78 & 78.88 & 77.03 & 89.76 & 90.99 &  93.93 & $\cdot$ & $\checkmark$    \\
 & 81.10 & 79.92 & 77.79 & 89.88 & 91.07 &  94.09 & $\checkmark$ & $\checkmark$    \\
\bottomrule[1.5pt]
\end{tabular}} \vskip-0.1in
\end{table*}

\subsection{Sensitivity Analysis of Parameters}
We investigate the effect of different hyperparameters, i.e., preserving (dropping) edge ratio, the balance parameter $\alpha$, and the judgments of views $\{g_n^m\}_{m=1}^M$, where $M$ is set to $2$.

\subsubsection{Sensitivity to the degree of perturbation}
\begin{figure}[!t]
    \centering
    \includegraphics[width=0.50\textwidth]{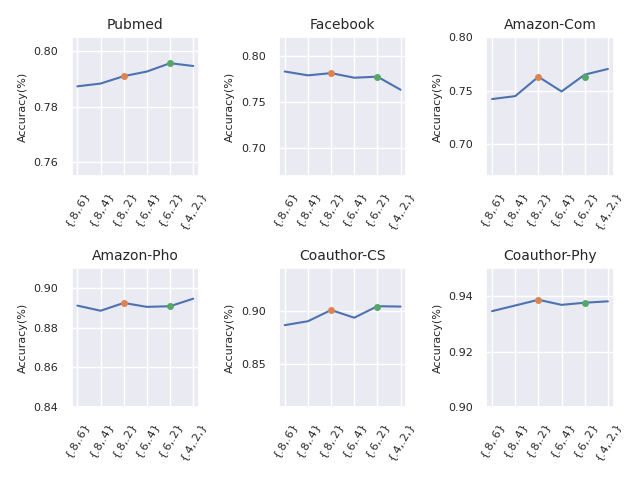}\vskip-0.1in
    \caption{Performance vs. different perturbations. Contrastive learning on graphs is performed with various degrees of perturbations. The orange dot and the green dot denote the performances of two views $\{0.8,0.2\}$ and $\{0.6,0.2\}$. }
    \label{view} \vskip-0.1in
\end{figure}

To choose ordered views for C2F, we analyze the effect of different preserving edge ratios in the vanilla CL, where we set $\alpha$ to $1$ and  the judgments of pair views to $\{1,1\}$. First, we investigate the performance of the different view compositions generated from different preserving edge ratios and report the classification results on all datasets in Fig.~\ref{view}. To do this, we vary the preserving edge ratio in the range of $[0.8,0.6,0.4,0.2]$ and construct six compositions of pairwise views, i.e., $\{0.8,0.6\}$, $\{0.8,0.4\}$, $\{0.8,0.2\}$, $\{0.6,0.4\}$, $\{0.6,0.2\}$ and~$\{0.4,0.2\}$. 

Fig.~\ref{view} shows that the performance is sensitive to the dropping edge ratio. The underlying trend of performance is still upward with the decrease of preserving edge ratios on all datasets except for Facebook. We assume that a higher degree of perturbation can enhance pre-training encoder robustness, which can increase the performance of downstream tasks. Besides, we find that the accuracy on both $\{0.8,0.2\}$ (i.e., the orange dot) and $\{0.6,0.2\}$ (i.e., the green dot) is relatively higher than other view compositions on all datasets. Specially, though sometimes the model achieves better results on $\{0.4,0.2\}$, we choose $\{0.6,0.2\}$ as the preserving edge ratios of two views, which makes the order of views more obvious.

\subsubsection{Sensitivity to Balance Parameter}
\begin{figure}[!t]
    \centering
    \includegraphics[width=0.50\textwidth]{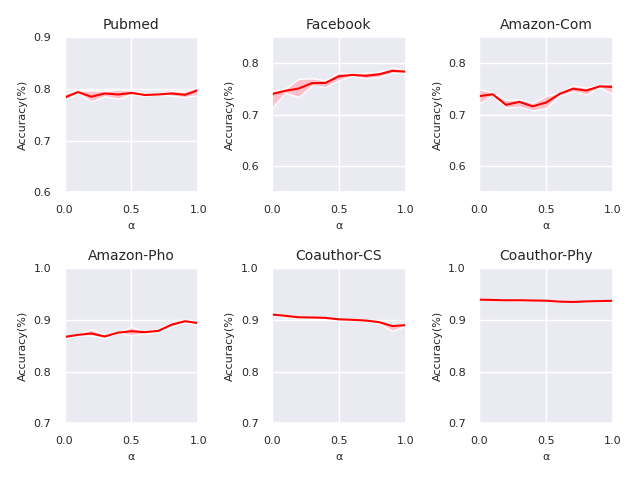}\vskip-0.1in
    \caption{Accuracy vs. the balance parameter $\alpha$. The fine-grained ranking module is performed with various $\alpha$ on all datasets. The shaded part represents the variance of the five runs.}
    \label{lambda} \vskip-0.1in
\end{figure}

We further investigate the impact of our method on the balance parameter $\alpha$. We conduct experiments to perform fine-grained ranking with different values of $\alpha$ (i.e., $0.1, 0.2, 0.3, 0.4, 0.5, 0.6, 0.7, 0.8, 0.9, 1$), where the judgments of coarse ranking for two views are set to \{1,1\}. The results on all datasets are plotted in Fig.~\ref{lambda}. In particular, the classification performance of our method increases lightly with the $\alpha$. Obviously, Pubmed, Coauthor-CS, and Coauthor-Phy are not sensitive to $\alpha$. That means the weight of fine-grained ranking is higher on those datasets. We assume that the densities of those three datasets are all very low, so graph structures of those datasets are easier to be destroyed when edges are removed randomly. So the fine-grained ranking is more important to prevent the reduction of node discrimination in the sparse graph. What's more, for Facebook, Amazon-Com, and Amazon-Pho, the best performance can be achieved when $\lambda=0.9$. 

\subsubsection{Sensitivity to Judgments}
We perform our coarse ranking's sensitivity to the judgments on Coauthor-Phy and Amazon-Com datasets, where the preserving edge ratio of two views are both set to $\{0.6,0.2\}$, and the balance parameter $\alpha$ is set to $0.9$. We fix the judgment of one view as $1$ and vary the judgment of another view within $[0.2, 0.4, 0.6, 0.8, 1]$. The two judgments need to be normalized via the softmax function $\sigma(x_i) = \frac{\exp(x_i)}{\sum_{k=1}^2 \exp(x_k)}$, $\forall x_1, x_2 \in [0.2, 0.4, 0.6, 0.8, 1]$. For any components of $\{x_1,x_2\}$ that satisfy $x_1-x_2=d$ and $d$ is a constant, $\sigma(x_1)$ and $\sigma(x_2)$ are fixed. So the performances are equal on the diagonal of the heat map.

\begin{figure}[!t]
    \centering
    \includegraphics[width=0.35\textwidth]{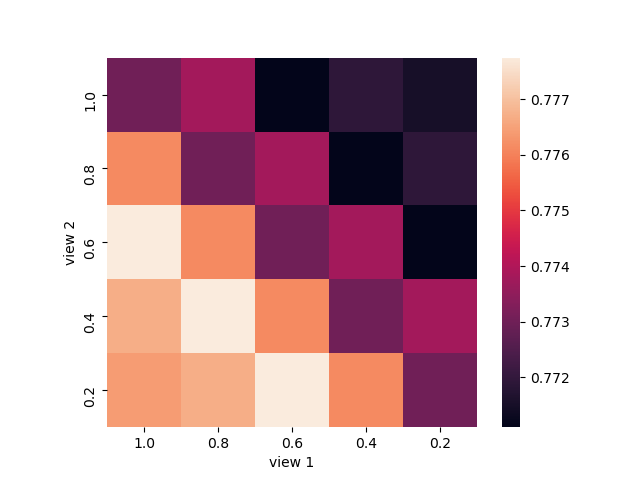}\vskip-0.1in
    \caption{The sensitivity to judgment in Amazon-Com.}\vskip-0.1in
    \label{fig:judgment_com}
\end{figure}

Fig.~\ref{fig:judgment_com} shows the test accuracy using different judgment values for two views on the Amazon-Com dataset. When the judgment of view one (i.e., preserving edge ratio=0.6) is larger than the judgment of view two (i.e., preserving edge ratio=0.2), C2F achieves better performance (the lower left part of the heat map is lighter than the upper right part), which illustrates the view 1 is more important than view 2 in C2F. Besides, when the judgment of view 2 is equal to 0.6, C2F achieves the top performance, which implies that coarse ranking can capture the order among two degrees of perturbations correctly according to the similarity between views and the original graph. Furthermore, we can obverse that the performance of C2F decreases with the increase of the judgment of view 2.

\subsection{Comparison with Baselines on Image Dataset}~\label{image1}

We further confirm the effectiveness of C2F model on an image dataset. The algorithm~\ref{C2F on images} of C2F on images is given in APP.~\ref{the pieline of C2F on images}. We pretrain a ResNet encoder by minimizing the C2F loss and obtaining self-supervised image representations for the downstream image classification task.

We follow the experiment settings of SimCLR~\cite{DBLP:conf/icml/ChenK0H20} on CIFAR-10. We use ResNet-50 as the base encoder. We set the batch size to be 128, the learning rate is set to 1.0 and the temperature parameter is set to 0.5. We adopt color augmentation with strength varying from 0 to 1. In our experiments, the strengths of two views are set to $\{0.8, 0.4\}$, and those judgments of two views are set to $\{0.6, 1\}$. The balance parameter $\alpha$ is set to $0.8$. The experiment results of sensitivity analysis on hyperparameter $\alpha$ and the judgments of two views are reported in Table~\ref{alphacifar} and Table~\ref{judgmentcifar} in APP.~\ref{B}.

\begin{figure}[!t]
    \centering
    \includegraphics[width=0.38\textwidth]{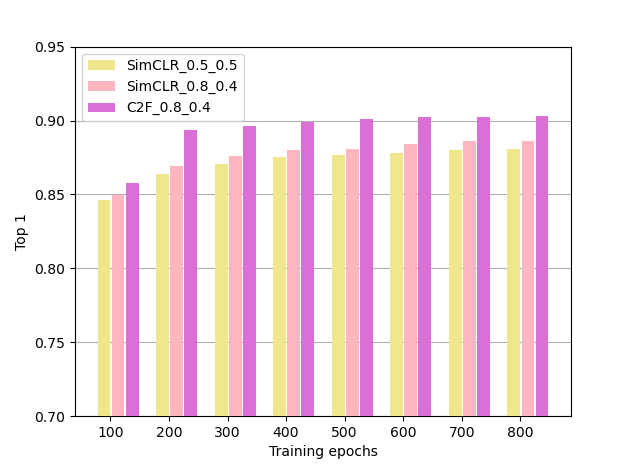}\vskip-0.1in
    \caption{ResNet-50 trained with different methods and epochs. The accuracy(\%) on each bar is the average of three runs.}\vskip-0.1in
    \label{cif}
\end{figure}

Fig.~\ref{cif} shows the performance of the approaches trained with different numbers of epochs. ``SimCLR\_0.5\_0.5" and ``SimCLR\_0.8\_0.4" represent SimCLR with different color strengths of \{0.5, 0.5\} and \{0.8, 0.4\} respectively for two views. ``C2F\_0.8\_0.4" denotes our C2F on images with color strengths of \{0.8, 0.4\}. ``SimCLR\_0.8\_0.4" performs better than ``SimCLR\_0.5\_0.5" for all epochs, indicating that the difference in perturbation degrees has an impact on the robustness of the encoder. C2F obtains an absolute improvement of approximately 1.71\% than SimCLR\_0.8\_0.4 on top 1 accuracy\footnote{Top 1 accuracy is the conventional accuracy of the top 1 prediction on classification tasks.} and improves the accuracy by 2\% compared to SimCLR\_0.5\_0.5, which further demonstrates the effectiveness of our proposed C2F model.

\section{Conclusion}\
\label{Conclusion}
In this paper, we analyze that contrastive learning is a special case of learning to rank, which ranks the positive sample at the top and ranks negative samples indiscriminately at the bottom. To incorporate the order relation in multi-positive/negative samples, we propose a generalized ranking framework, Coarse-to-Fine Contrastive learning, unifying the proposed coarse ranking and fine-grained ranking models. The coarse ranking model assigns ordered judgments for positive samples, which can encourage the encoder to utilize the prior information among ordered positive views. And the fine-grained ranking module allows negative samples to participate in sorting via assigning self-generated ground truth for negative samples, which can avoid the information loss caused by the strong augmentations. The experimental results over graph datasets manifest the effectiveness of C2F to discriminative information among positive and negative samples for enhancing graph representation learning. 

We have conducted experiments of C2F on the image classification task to demonstrate the potential of our C2F in computer vision (CV). Moreover, C2F can benefit other artificial intelligence fields, such as natural language processing (NLP), and imitation learning (IL), thus paving a foundation for further studies of the generalization of C2F. In general, the order relation between augmented views can be inferred from pre-defined ordered augmentations, but the difference for diverse fields lies in the specific augmentation strategies.


\normalem

\bibliographystyle{IEEEtran}

\newpage

\begin{IEEEbiography}[{\includegraphics[width=1in,height=1.25in,clip,keepaspectratio]{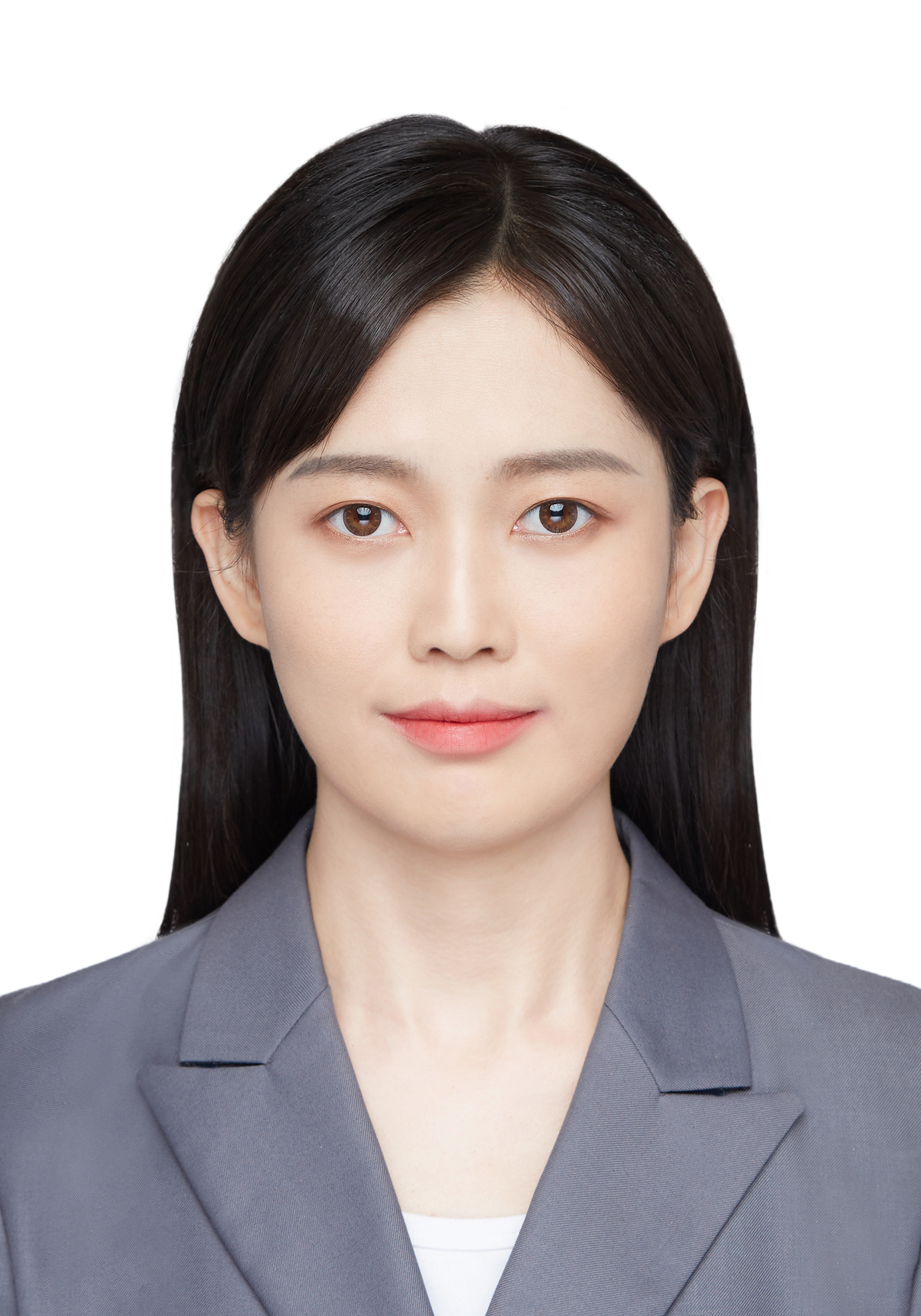}}]{Peiyao Zhao} received MSc degree in School of Mathematics and Statistics from Beijing Institute of Technology (BIT), China. She currently interns at A*STAR Center for Frontier AI Research in Singapore and is pursuing a Ph.D. degree from BIT.  Her research focuses on Contrastive Learning, Graph Representation Learning.
\end{IEEEbiography}

\begin{IEEEbiography}[{\includegraphics[width=1in,height=1.25in,clip,keepaspectratio]{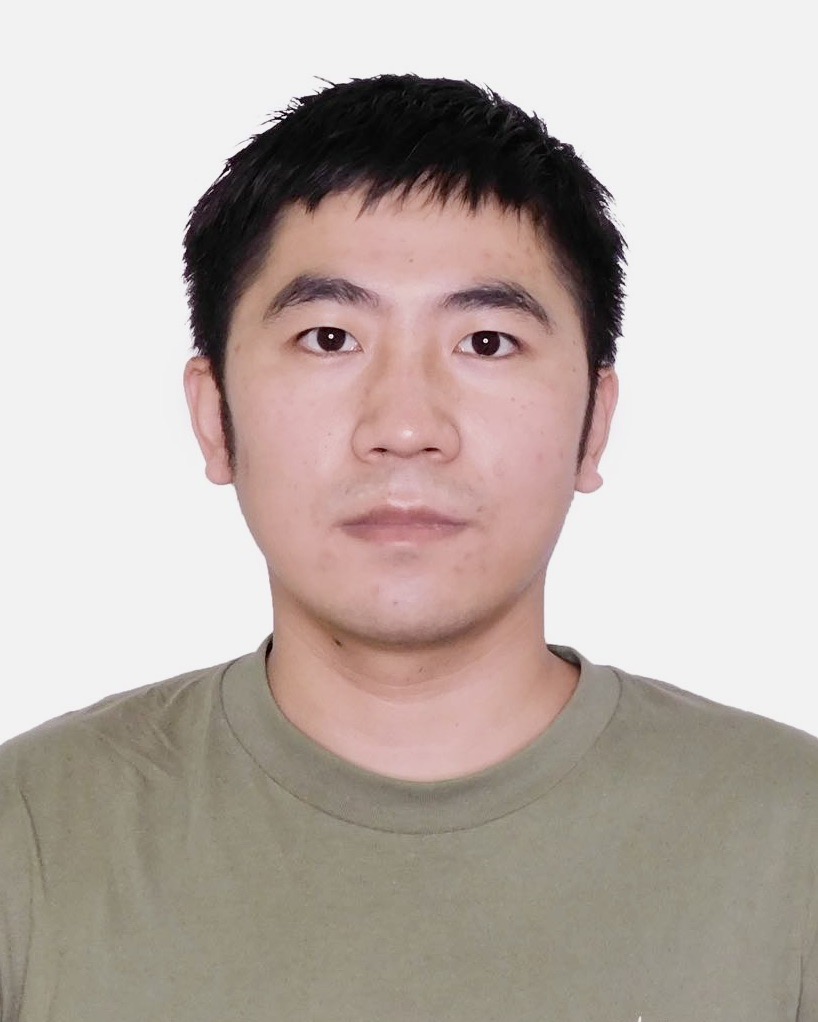}}]{Yuangang Pan}
is working as a research scientist at A*STAR Centre for Frontier AI Research. He completed his Ph.D. degree in Computer Science in Mar 2020 from University of Technology Sydney (UTS), Australia. Before joining A*STAR, he was a postdoctoral research associate at the Australian Artificial Intelligence Institute at UTS. He has authored or co-authored papers on various top conferences and journals, such as AAAI, IEEE TIFS, IEEE TKDE, IEEE TNNLS, ACM TOIS, MLJ, and JMLR. His research interests include Deep Clustering, Deep Generative Learning, Differential Privacy, and Robust Ranking Aggregation.
\end{IEEEbiography}

\begin{IEEEbiography}[{\includegraphics[width=1in,height=1.25in,clip,keepaspectratio]{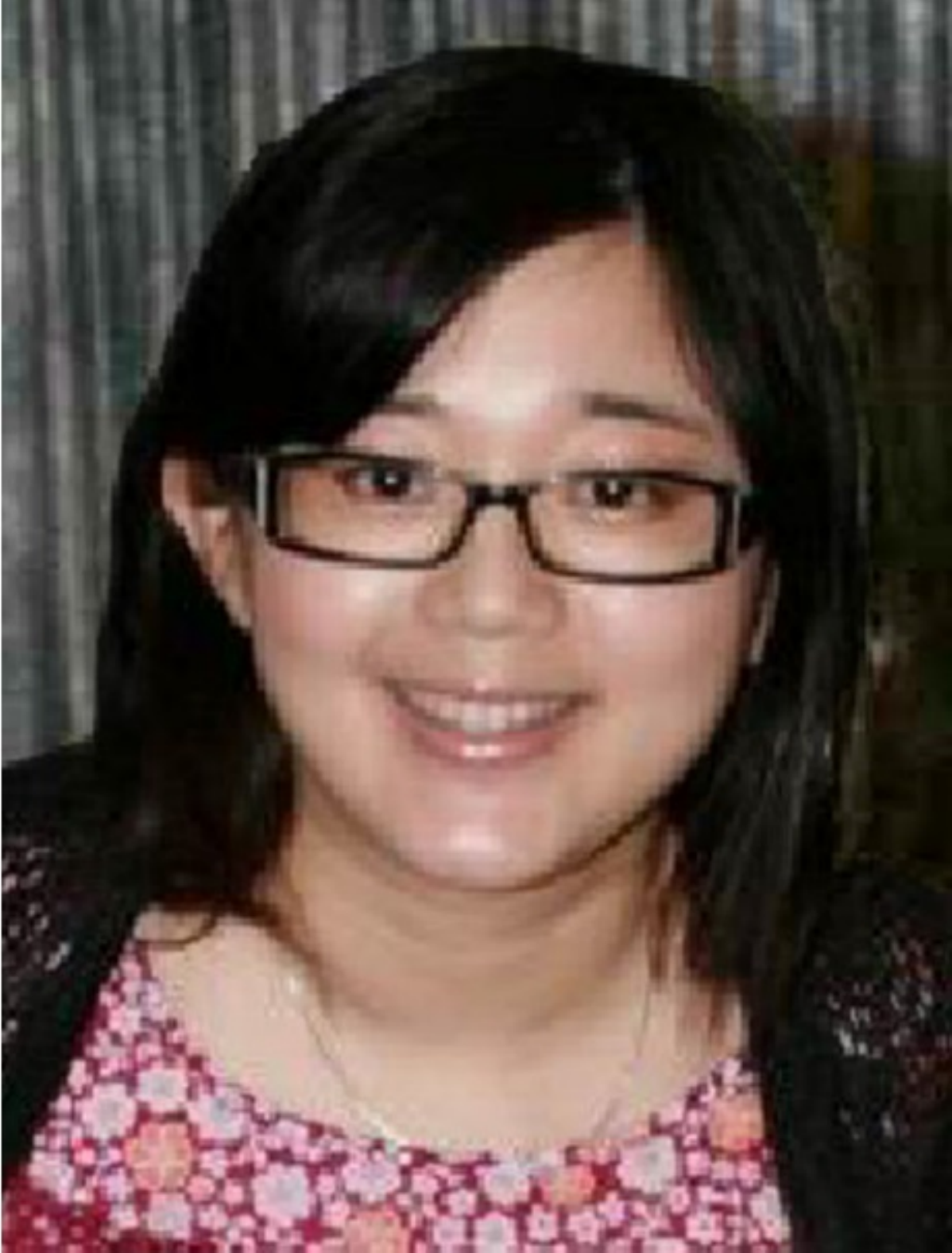}}]{Xin Li}
is currently an Associate Professor in the School of Computer Science at Beijing Institute of Technology, China. She received the B.Sc. and M.Sc degrees in Computer Science from Jilin University China, and the Ph.D. degree in Computer Science at Hong Kong Baptist University. Her research focuses on the development of algorithms for representation learning, reasoning under uncertainty, and machine learning with application to Natural Language Processing, Recommender Systems, and Robotics.
\end{IEEEbiography}
\begin{IEEEbiography}[{\includegraphics[width=1in,height=1.25in,clip,keepaspectratio]{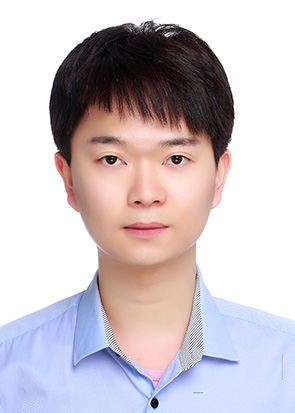}}]{Xu Chen}
received a Ph.D. degree at Cooperative Meidianet Innovation Center in Shanghai Jiao Tong University in 2021. He also received a dual Ph.D. degree at University of Technology, Sydney in 2021. He has published papers of top conferences and journals such as CVPR, IEEE TPAMI and ACM TOIS. He is now working at Alibaba Group. His research interests include machine learning, graph representation learning, self-supervised learning and recommendation systems.
\end{IEEEbiography}
\begin{IEEEbiography}[{\includegraphics[width=1in,height=1.25in,clip,keepaspectratio]{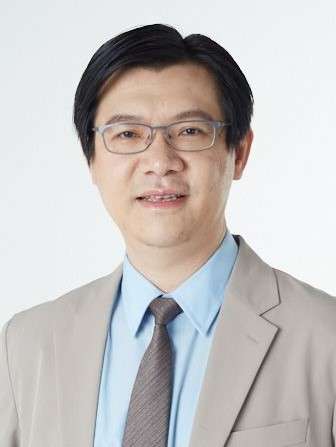}}]{Ivor W. Tsang} (Fellow IEEE) is currently the Director of the A*STAR Centre for Frontier AI Research, Singapore. He is also a Professor of artificial intelligence with the University of Technology Sydney, Ultimo, NSW, Australia, and the Research Director of the Australian Artificial Intelligence Institute. His research interests include transfer learning, deep generative models, learning with weakly supervision, Big Data analytics for data with extremely high dimensions in features, samples and labels.

In 2013, he was the recipient of the ARC Future Fellowship for his outstanding research on Big Data analytics and large-scale machine learning. In 2019, his JMLR paper Towards ultrahigh dimensional feature selection for Big Data was the recipient of the International Consortium of Chinese Mathematicians Best Paper Award. In 2020, he was recognized as the AI 2000 AAAI/IJCAI Most Influential Scholar in Australia for his outstanding contributions to the field between 2009 and 2019. Recently, he was conferred the IEEE Fellow for his outstanding contributions to large-scale machine learning and transfer learning. He serves as the Editorial Board for the JMLR, MLJ, JAIR, IEEE TPAMI, IEEE TAI, IEEE TBD, and IEEE TETCI. He serves/served as a AC or Senior AC for NeurIPS, ICML, AAAI and IJCAI, and the steering committee of ACML.
\end{IEEEbiography}
\begin{IEEEbiography}[{\includegraphics[width=1in,height=1.25in,clip,keepaspectratio]{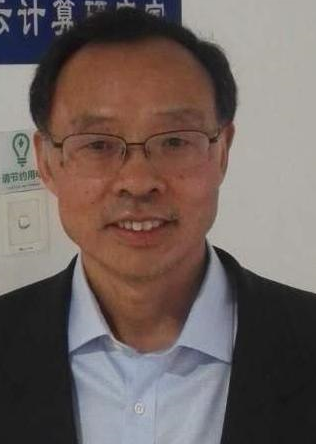}}]{LejianLiao} received the Ph.D. degree from the Institute of Computing Technology, Chinese Academy of Sciences, in 1994. He is currently a Professor with the School of Computer Science and Technology, Beijing Institute of Technology. He has published numerous papers in several areas of computer science. His research interests include machine learning, natural language processing, and intelligent networks.

\end{IEEEbiography}

\section{Appendix}
\appendices

\subsection{The Summarization of Related Work}\label{The summarization of related work}
Tab.~\ref{methods comparsion} provides a brief comparison between our C2F and the related graph contrastive-based methods, where ``Ordered views" denotes whether the method incorporates the order relationship of different perturbation degrees in augmented views. Roughly, existing works can be divided into two categories. The difference between them lies in whether the negative samples are required in their respective framework.
\begin{table}[htbp!]
\centering
\caption{\label{methods comparsion} The summarization of baselines.}
\setlength{\tabcolsep}{1.2mm}
\scalebox{0.75}{
\begin{tabular}{ccccccc}\toprule[1.5pt]
Paradigm & Method & Classification task& \#Augmentation type     & Ordered views\\\midrule[1.0pt]
& DGI & Node & 1  & No \\
&InfoGraph & Graph & 1  & No \\
&GCC & Node & 1  & No \\
Neg. Samples Required &GraphCL & Graph & $>1$  & No \\
&MVGRL & Node\&Graph & $>1$  & No \\
& GCA & Node & $>1$  & No \\
&\textbf{Our C2F} & Node & $>1$  & \textbf{Yes}\\
\hline
Neg. Samples NOT Required & BGRL & Node & -   & No \\
&G-BT & Node &  -   & No \\
\hline
\bottomrule[1.5pt]
\end{tabular}}
\vskip-0.1in  \end{table}

\subsection{Some Insights for Contrastive Learning}~\label{insights}

\begin{figure}[!t]
    \centering
    \includegraphics[width=0.35\textwidth]{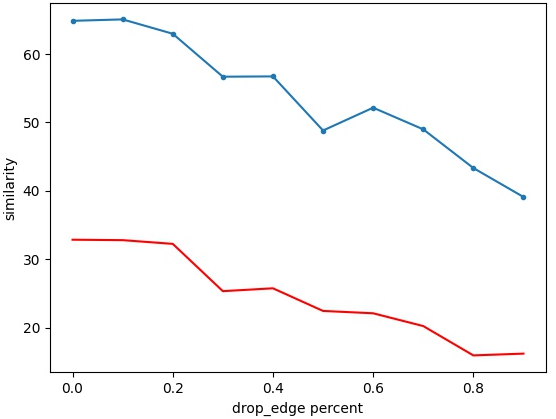}\vskip-0.1in 
    \caption{The effect of drop edge ratio. The blue line depicts that the average node similarity between one view and the original graph decreases with the increase of dropping edge rate of the view. The red line shows that all node similarity on average inside one view also decreases with the increase of dropping edge ratio of the view.} \vskip-0.1in
    \label{fig:similarity}
\end{figure}

The stronger the perturbation is, the less similar the augmented view and original graph are. To illustrate the correlation between the drop rates and the graph similarity, we adopt GCNs to perform the node classification on the graphs (the original one and augmented ones) of a benchmark data set (i.e., Pubmed) and obtain the embedding of nodes for each graph. Fig.~\ref{fig:similarity} depicts the correlations between the drop rate and the average node similarity for all positive pairs. The blue line shows that the average node similarity between views of different magnitudes and the original graph decreases with the increase of drop rate, which supports the motivation of the coarse ranking model.

Moreover, the red line in Fig.~\ref{fig:similarity} shows that all node similarity on average inside one view also decreases with the degree of perturbation added to the view. Namely, the discrimination of nodes inside one view will reduce because GCN leverages less graph structure in the augmented view than in the original graph, which is not conducive to learning discriminative node representation. To prevent the loss of structure information, we propose the fine-grained ranking loss that allows negative samples to participate in sorting. The supervision of negative samples can encourage the encoder to learn discriminative node representations, namely, avoiding discrimination reduction of all nodes.

\subsection{Dataset Splitting on Graph Datasets}~\label{split}
We note that the dataset split on the node classification can influence the performance of graph CL methods in the evaluation stage. The way of data split used in many graph CL works (i.e., DGI~\cite{DBLP:conf/iclr/VelickovicFHLBH19},  MVGRL~\cite{DBLP:conf/icml/HassaniA20},GCC~\cite{DBLP:conf/kdd/QiuCDZYDWT20}) originates from the semi-supervised works of graph representation learning (i.e, GCN~\cite{DBLP:conf/iclr/KipfW17}, GAT~\cite{DBLP:conf/iclr/VelickovicCCRLB18}, GraphSage~\cite{DBLP:conf/nips/HamiltonYL17}). Whereas some graph CL works, e.g., (GRACE~\cite{zhu2020deep}, BGRL~\cite{thakoor2021bootstrapped}, G-BT~\cite{DBLP:journals/corr/abs-2106-02466}), utilize a random split of the nodes into $(10\%/10\%/80\%)$ training/validation/test set. Usually, such a way of split results in a larger train/validation set than that in the former way and favors the training, except for Coauthor-Phy. In this paper, we follow the former way of data split as in the semi-supervised setting to evaluate the performance of our method on the downstream task. 
For Pubmed, we follow the common data splits in GCN~\cite{DBLP:conf/iclr/KipfW17} and DGI~\cite{DBLP:conf/iclr/VelickovicFHLBH19}. For Facebook, Coauthor-CS, Amazon-Com and Amazon-Pho, we follow the setting in~\cite{DBLP:journals/corr/abs-1811-05868}, where 20 nodes of each class are randomly sampled as the train set and 30 nodes of each class are randomly sampled as the validation set and the rest is the test set.
For Coauthor-Phy, we follow the data split of the inductive learning setting in many works. That is, the test nodes are not seen during training. We consider training models on graphs following the former data split of the semi-supervised setting is more challenging. The details of two data splits on the benchmark datasets are summarized in Table~\ref{data split}.

\begin{table}[H]
\centering
\caption{\label{data split} Comparison of dataset split for evaluation. In the train/validation/test lines, \textbf{bold} numbers denote the data split of some baselines and normal numbers denote the data split of our C2F.}
\setlength{\tabcolsep}{1.2mm}
\scalebox{0.75}{
\begin{tabular}{c|cccccc}\toprule[1.5pt]
 \diagbox{Split}{Dataset} & Pubmed & Facebook & Amazon-Com & Amazon-Pho & Coauthor-CS & Coauthor-Phy \\\midrule[0.8pt]
Total nodes & 19717 & 22470 & 13752 & 7650 & 18333 & 34493 \\
\hline 
\multirow{2}*{Train}& 60 & 80 & 200 & 160 & 300 & 20000\\
~ & \textbf{\bf{1971}} & \textbf{2247} & \textbf{1375} & \textbf{756} & \textbf{1834} & \textbf{3450}\\
\hline \multirow{2}*{Validation}& 500 & 120 & 300 & 240 & 450 & 5000\\
~ & \textbf{1972} & \textbf{2247} & \textbf{1375} & \textbf{756} & \textbf{1833} & \textbf{3449}\\
\hline
\multirow{2}*{Test}& 1000 & 22270 & 13252 & 7250 & 17583 & 9493 \\
~ & \textbf{15774} & \textbf{17976} & \textbf{11002} & \textbf{6120} & \textbf{14666} & \textbf{27594}   \\
\hline
\bottomrule[1.0pt]
\end{tabular}}
\end{table}

\subsection{Hyper-parameter Setting on Graph Datasets}~\label{Hyper-parameter}
Following the evaluation method in many works (i.e. DGI~\cite{DBLP:conf/iclr/VelickovicFHLBH19}, GCC~\cite{DBLP:conf/kdd/QiuCDZYDWT20}, MVGRL~\cite{DBLP:conf/icml/HassaniA20}), the learned encoder needs to be fixed and we train a one-layer linear classifier for evaluation. Trainable parameters of all models are initialized with uniform distribution. The best-trained classifier is chosen according to the performance on the validation set. For Amazon-Com, Amazon-Pho, the weight decay is set to 0.0. For other benchmarks, it is set to 5e-4. All reported results are averaged over five independent runs under the same configuration.
\subsection{Additional Metrics on Node Classification}~\label{metrics}
Following the evaluation protocol of CGNN~\cite{DBLP:journals/corr/abs-2008-11416} and its data split, we have compared C2F with 3 state-of-the-art graph CL methods, GCA~\cite{DBLP:conf/www/0001XYLWW21}, BGRL~\cite{thakoor2021bootstrapped}, and G-BT~\cite{DBLP:journals/corr/abs-2106-02466}, on additional metrics, F1-Score, AUC, and Recall for six benchmark datasets (see Table~\ref{additional metrics}). Please note that GCA is considered one of the most state-of-the-art Graph CL approaches requiring negative samples in their paradigms. BGRL and G-BT are the two state-of-the-art Graph CL approaches in which negative samples are not required. 

\begin{table}[H]
\centering
\caption{\label{additional metrics} Performance of different models on F1, AUC and Recall score (\%).}
\setlength{\tabcolsep}{1.2mm}
\scalebox{0.75}{
\begin{tabular}{cccccccc}\toprule[1.5pt]
 Metric & Method & Pubmed & Facebook & Amazon-Com & Amazon-Pho & Coauthor-CS & Coauthor-Phy \\\midrule[0.8pt]
 \multirow{5}{*}{F1}
 ~ & GCA~\cite{DBLP:conf/www/0001XYLWW21} & 80.19 & 63.54 & 72.19 & 83.57 & 85.70 & 64.71 \\
  ~ & BGRL~\cite{thakoor2021bootstrapped} & 59.75 & 59.28  & \textbf{76.44} & 85.32 & 86.27 & 67.65 \\
  ~ & G-BT~\cite{DBLP:journals/corr/abs-2106-02466}  & 80.06  & 67.08 &  76.00 & 81.79 & 85.88 & 66.87 \\
  ~ & {\bf C2F}  & \bf{80.32} & \bf{78.53} & 75.69 &  \bf{88.99} & \bf{86.39} & \bf{90.99} \\
  \midrule[0.8pt]
 \multirow{5}{*}{AUC}
 ~ & GCA~\cite{DBLP:conf/www/0001XYLWW21} & 92.45 & 82.17 & 97.15 & 97.78 & 89.95 & 89.39\\
  ~ & BGRL~\cite{thakoor2021bootstrapped} & 77.10 & 78.81  & 97.30 & 98.04 & 98.90 & 93.10 \\
  ~ & G-BT~\cite{DBLP:journals/corr/abs-2106-02466} & 91.31 & 86.35 & 95.24 & 97.28 & 98.53 & 92.03 \\
  ~ & {\bf C2F}  & \bf{93.03} & \bf{93.70} & 96.72 &  \bf{98.71} & \bf{99.00} & \bf{99.00} \\
  \midrule[0.8pt]
   \multirow{5}{*}{Recall}
 ~ & GCA~\cite{DBLP:conf/www/0001XYLWW21} & 80.11 & 67.52 & 81.41 & 87.86 & 87.10 & 57.06 \\
  ~ & BGRL~\cite{thakoor2021bootstrapped} & 63.84 & 59.63  & \textbf{85.71} & 89.37 & 87.05 & 62.58 \\
  ~ & G-BT~\cite{DBLP:journals/corr/abs-2106-02466} & 80.50  & 73.09 & 82.12 & 87.73 & 87.20 & 60.76 \\
  ~ & {\bf C2F}  & \bf{81.00} & \bf{79.32} & 84.55 &  \bf{91.08} & \bf{87.62} & \bf{89.76} \\
\hline
\bottomrule[1.0pt]
\end{tabular}}
\end{table}
We can observe from Table~\ref{additional metrics} that, our C2F achieves on par or better performance in terms of all metrics on all datasets than those baselines, which further verifies the effectiveness of incorporating the prior of the order relation between positive/negative samples into the encoder training. C2F is inferior to BGRL and GCA on Amazon-Com, which may be because C2F suffers from an increasing number of false negative samples on denser graphs. Amazon-Com is a relatively dense graph with the density $0.13\%$ (see Table II). C2F learns more similar node representations on Amazon-Com, thus drawing more false negative samples to deteriorate C2F performance, but BGRL doesn't require negative samples. Whereas GCA designs an adaptive augmentation strategy to capture more neighbor information to improve performance.

\subsection{The Analysis of Coarse-to-Fine Ranking}~\label{The analyze of Coarse-to-Fine ranking}
Here we prove that $J^a_n$ is a normalized judgment probability matrix, namely 
\begin{align}
&\sum_{m=1}^M \sum_{k=0}^K (J^a_n)_{mk} \\
&= \alpha \sum_{m=1}^M \sum_{k=0}^K (J^c_n)_{mk} + (1-\alpha)\sum_{m=1}^M \sum_{k=0}^K (J^f_n)_{mk} \nonumber\\
& = \alpha \sigma(\textbf{g}^c_{1:M}) \ast \textbf{e}^T + (1-\alpha) \frac{1}{M}\ast \mathbf{1} \ast \sigma(\textbf{g}_n)^T\nonumber\\
&= \alpha \sum_{m=1}^M \sigma(\textbf{g}^c_{1:M})_m + (1-\alpha)\ast \frac{1}{M}\sum_{m=1}^M \sum_{k=0}^K \sigma(\textbf{g}_n)_k \nonumber\\
& = \alpha + (1-\alpha) \ast \frac{1}{M} \ast M\nonumber\\
& = 1.\nonumber
\end{align}

In Eq.~\eqref{overall_loss}, when $M=1$ and $\lambda=0$, it is contrastive learning with only one view, which only maximizes the agreement of positive samples and ignores the difference of negative samples. When $M\geqslant2$ and $\lambda > 0$, it is our Coarse-to-Fine ranking model, which encourages the encoder to capture discriminative information among ordered augmented views and negative samples. That is beneficial for the encoder to learn distinguished node representation via incorporating those prior knowledge.
\subsection{The Framework of C2F on Images}~\label{the pieline of C2F on images}
The differences between C2F on images and that on graphs lie in the following three aspects shown in Tab.~\ref{differences}: 

\begin{table*}[htbp!]
\centering
\caption{\label{differences} The differences between C2F on graphs and C2F on images. ''Neg. Samples" denotes negative samples.}
\setlength{\tabcolsep}{1.2mm}
\scalebox{0.9}{
\begin{tabular}{c|c|c|c}\toprule[1.5pt]
 & Augmentation & Encoder & Neg. Samples \\ \hline
C2F on graphs & Edge dropping/Feature masking & GAT & Drawn in the given graph   \\
\hline
C2F on images & Color distortion & ResNet-50 & Other samples' views in the batch  \\
\bottomrule[1.5pt]
\end{tabular}}
\vskip-0.1in  \end{table*}

The augmentation strategy of C2F on images is color distortion, where we fix the color dropping and only vary the degree of color jittering as the strength hyperparameter (including brightness, contrast, saturation, and hue). The negative samples in C2F on graphs are randomly drawn in the original graph but those in C2F on images are other images' augmented views in the batch. Rather than the above differences, the pairs generation and the C2F loss for C2F on graphs and images are the same. We present the algorithm of C2F on images in Algorithm 1 and its framework comprises the following four major components.

\begin{itemize}
    \item Data augmentation. For each image $\boldsymbol{x}_i$ in minibatch, a list of correlated augmented views can be constructed via varying the color strength $\theta$, namely $\mathcal{X}_i=\{\boldsymbol{x}_i^{m}|\boldsymbol{x}_i^{m}=\mathcal{T}_{\theta_m}(\cdot|\boldsymbol{x}_i),m=1,\dots,M\}$, where $\{\theta_m\}_{m=1}^M$ satisfies the constraint $\theta_1>\ldots>\theta_M$.
    
    \item Feature extraction. Then they go through a shared encoder $f_\theta$, resulting in a list of embedded representations $\mathcal{Z}_i=\{\boldsymbol{z}_i^m\}^{M}_{m=1}$ sorted by the degree of the color perturbation for the image $\boldsymbol{x}_i$.
    \item Pairs generation. In C2F model, each image $\boldsymbol{x}_i$ has $M\times(K+1)$ pairs of samples, including $M$ positive sample pairs $\{(\boldsymbol{z}_i,\boldsymbol{z}_i^m)\}_{m=1}^M$ and $M\times K$ negative sample pairs $\{(\boldsymbol{z}_i^m,\boldsymbol{z}_i^k)\}_{k=1}^K, m=1,\dots,M$. 
    \item The C2F loss. The predicted ranking score is defined as the normalized probability of each pair's similarity over all pairs' in Eq.(9). While the ground truth scores are defined as the weighted judgment probability matrix from the coarse and fine-grained ranking model in Eq.(14). Finally, Eq.(13) is the C2F loss on image classification task same as that on graph classification.
\end{itemize}

\begin{algorithm}[!t]
\begin{algorithmic}[1]
\caption{\label{C2F on images}coarse-to-fine contrastive learning on images.}
\STATE \textbf{Input}: batch size $N$, augmentation parameters $\theta_1>\ldots>\theta_M$, structure of $\mathcal{T},f$, similarity function $s(\cdot,\cdot)$, hyperparameter $\alpha$.\\
	\FOR{sampled minibatch $\{\boldsymbol{x}_i\}_{i=1}^N$}
	\FOR{{\bf{all}} $i\in\{1,\dots,N\}$}
	\STATE $\boldsymbol{z}_i=f(\boldsymbol{x}_i)$ \quad{\color{gray}\# apply encoder}\\
	\STATE $\boldsymbol{x}_i^m = \mathcal{T}_{\theta^m}(\boldsymbol{x}_i)$, $m = 1,2,\ldots,M.\quad $ {\color{gray}{\# create M views}}\\
	\STATE $\boldsymbol{z}_i^m = f(\boldsymbol{x}_i^m)$, $m = 1,2,\ldots,M.\quad ${\color{gray}{\# feature extraction}}
	\ENDFOR
	\FOR{\textbf{all} $i\in\{1,\dots,N\}$}
	\STATE $s_{i,i}^m=\boldsymbol{z}_i^{T}\boldsymbol{z}_i^m/(\|\boldsymbol{z}_i\| \|\boldsymbol{z}_i^m\|), m=1,2,\dots,M$ {\color{gray} \# pairwise similarity}\\
	\STATE $s_{i,k}^{m}=\boldsymbol{z}_k^{T}\boldsymbol{z}_i^m/(\|\boldsymbol{z}_k\| \|\boldsymbol{z}_i^m\|), m=1,2,\dots,M, k=1,2,\dots,K$\\
	\STATE calculate the predicted ranking scores following Eq.(9)
	\STATE $s_{i,i}=\boldsymbol{z}_i^{T}\boldsymbol{z}_i/(\|\boldsymbol{z}_i\| \|\boldsymbol{z}_i\|)$
	\STATE $s_{i,k}=\boldsymbol{z}_k^{T}\boldsymbol{z}_i/(\|\boldsymbol{z}_k\| \|\boldsymbol{z}_i\|), k=1,2,\dots,K$
	\STATE calculate the ground truth ranking scores following Eq.(14)
	\ENDFOR

	\STATE{calculate the C2F ranking loss $\mathcal{L}_{CF}$ following Eq.(13)}
	\STATE{update network $f$ to minimize $\mathcal{L}_{CF}$}
	\ENDFOR
\STATE \textbf{return} the encoder network $f(\cdot)$
\end{algorithmic}
\end{algorithm}

\subsection{Ablation Study on Image Dataset}~\label{image2}
In this section, we perform ablation studies on the two components of C2F, including the coarse ranking and fine-grained ranking. To verify the effectiveness of the proposed C2F model, we further compare the variants of C2F to SimCLR on 800 epochs shown in Table.~\ref{Rank_9}. It can be seen that two downgraded models, including Coarse ranking and fine-grained ranking, both outperform SimCLR, which verifies the effectiveness of our coarse and fine-grained model. And our approach C2F which jointly applies Coarse and fine-grained ranking loss significantly outperforms two downgraded models which further shows the necessity of jointly considering two schemes.  
\begin{table}[htbp!]
\centering
\caption{\label{Rank_9} The accuracy(\%) of model variants along with SimCLR on CIFAR-10 in ablation study}
\setlength{\tabcolsep}{1.2mm}
\scalebox{1}{
\begin{tabular}{cc}\toprule[1.5pt]
 Methods & CIFAR-10  \\\midrule[1.0pt]
SimCLR & 88.10   \\
\hline
Coarse ranking & 88.60 \\
\hline
Fine-grained ranking & 90.29 \\
\hline
C2F & 90.31   \\
\bottomrule[1.5pt]
\end{tabular}}
\vskip-0.1in 
\vskip-0.1in  \end{table}

\subsection{Sensitivity to Balance Parameter on CIFAR-10}\label{B}
To choose appropriate $\alpha$, we pre-train the C2F ranking model with 200 epochs, where the color strengths are set to \{0.8,0.4\} and judgments of two views are set to \{1,1\}. The results are reported in Table~\ref{alphacifar}. We can get the best performance when $\alpha$ is set to 0.8. 
\begin{table}[!htbp]
  \caption{Performance vs. $\alpha$ on CIFAR-10.}
  \label{alphacifar}
  \centering
  \renewcommand{\arraystretch}{1.2}
  \setlength{\tabcolsep}{1.2mm}{	
  \scalebox{0.8}{
  \begin{tabular}{l|ccccccc}
    \toprule[1.3pt]      
    $\alpha$ & 0 & 0.2 &  0.4  & 0.6 & 0.8 & 1\\
    \midrule[1pt]
    Acc(\%) &  72.25 & 85.58 & 88.06 & 89.34 & 89.44 & 89.28\\
    \bottomrule[1.3pt]
  \end{tabular}}}
\vskip-0.1in 
\end{table}
\subsection{Sensitivity to Judgment on CIFAR-10}
To choose appropriate judgments for two views, we perform C2F ranking model with 200 epochs, where the color strengths are set to $\{0.8,0.4\}$ and $\alpha$ is set to 0.8, Similarly, we fix the judgment of one view (i.e., strength = 0.4) to 1, and adjust another judgment of another view (i.e., strength = 0.8) in the range of $[0, 0.2, 0.4, 0.6, 0.8, 1]$. The results are represented in Table~\ref{judgmentcifar}. We can get the best performance when the judgment of the second view $g^2$ is set to $0.6$. 
\begin{table}[!htbp]
  \caption{Performance vs. judgment on CIFAR-10.}
  \label{judgmentcifar}
  \centering
  \renewcommand{\arraystretch}{1.2}
  \setlength{\tabcolsep}{1.2mm}{	
  \scalebox{0.8}{
  \begin{tabular}{l|ccccccc}
    \toprule[1.3pt]      
    $g^2$ & 0 & 0.2 &  0.4  & 0.6 & 0.8 & 1\\
    \midrule[1pt]
    Acc(\%) &  86.70 & 88.27 & 88.96 & 89.39 & 89.10 & 89.05\\
    \bottomrule[1.3pt]
  \end{tabular}}}
\vskip-0.1in  
\end{table}


%





\ifCLASSOPTIONcaptionsoff
  \newpage
\fi

\end{document}